\documentclass[review]{elsarticle}

\usepackage{graphicx}
\usepackage{amsfonts}
\usepackage{amsmath,setspace,algorithm,algorithmic,multirow,cases,amssymb,subfigure}
\usepackage{lettrine}
\usepackage{xcolor}
\usepackage{times}
\usepackage{indentfirst}
\usepackage{booktabs}
\usepackage{blindtext}
\usepackage{array}
\usepackage{tabularx,booktabs}
\usepackage{chngpage}
\usepackage{float}
\usepackage{colortbl}
\usepackage{url}
\journal{Journal of \LaTeX\ Templates}

\begin{document}

\begin{frontmatter}

\title{A CNN-RNN Architecture for Multi-Label Weather Recognition}

\author[1]{Bin Zhao}
\author[2]{Xuelong Li}
\author[2]{Xiaoqiang Lu\corref{c}}
\cortext[c]{Corresponding author}
\ead{luxq666666@gmail.com}
\author[1]{Zhigang Wang}

\address[1]{School of Computer Science and Center for OPTical IMagery Analysis and Learning (OPTIMAL), Northwestern Polytechnical University, Xi'an 710072, Shaanxi, P. R. China.}
\address[2]{Xi'an Institute of Optics and Precision Mechanics,
	Chinese Academy of Sciences, \\Xi'an 710119, Shaanxi, P. R. China.}


\begin{abstract}
Weather Recognition plays an important role in our daily lives and many computer vision applications. However, recognizing the weather conditions from a single image remains challenging and has not been studied thoroughly. Generally, most previous works treat weather recognition as a single-label classification task, namely, determining whether an image belongs to a specific weather class or not. This treatment is not always appropriate, since more than one weather conditions may appear simultaneously in a single image. To address this problem, we make the first attempt to view weather recognition as a multi-label classification task, i.e., assigning an image more than one labels according to the displayed weather conditions. Specifically, a CNN-RNN based multi-label classification approach is proposed in this paper. The convolutional neural network (CNN) is extended with a channel-wise attention model to extract the most correlated visual features. The Recurrent Neural Network (RNN) further processes the features and excavates the dependencies among weather classes. Finally, the weather labels are predicted step by step. Besides, we construct two datasets for the weather recognition task and explore the relationships among different weather conditions. Experimental results demonstrate the superiority and effectiveness of the proposed approach. The new constructed datasets will be available at \url{https://github.com/wzgwzg/Multi-Label-Weather-Recognition}.
\end{abstract}


\end{frontmatter}


\section{Introduction} \label{intro}

The weather conditions influence our daily lives and production in many ways \cite{Lu:2014:TWC:2679600.2680161}, such as wearing, traveling, solar technologies and so on. Therefore, acquiring weather conditions automatically is important to a variety of human activities. A possible solution to weather recognition is utilizing various of hardwares. While these hardware equipments are usually expensive and need professionals to maintain. An alternative scheme is to recognize weather conditions from color images using computer vision techniques \cite{7784804,7351424}. Nowadays, surveillance cameras are ubiquitous, which makes the computer vision solution feasible. Apart from the guiding significance to our daily lives, weather recognition is also an important function to many other computer vision applications \cite{han2018advanced,han2018robust,cheng2018deep,han2017unified}, such as image retrieval \cite{qayyum2017medical}, image restoration \cite{6115972}, and the reliability improvement of outdoor surveillance systems \cite{7351424}. Robotic vision \cite{Shah:1997:MRN:283534.283535, 1249323} and vehicle assistant driving systems \cite{1505103, Yan:2009:WRB:1561386.1561431} can also benefit from the results of weather recognition. Thus, we can draw a simple conclusion that weather recognition from outdoor images has great research significance.

\subsection{Motivation and Overview}
Although weather recognition is of remarkable value, only a few researches have been published to tackle this problem. Several previous works \cite{1505103, Hautiére2006, 4621205, DBLP:conf/ivs/PavlicRI13} concentrated on recognizing weather conditions from images captured by in-vehicle cameras. Several other papers \cite{Song2014, 7294972, Lu:2014:TWC:2679600.2680161} exploited weather recognition from single outdoor images. All of these works referred to weather recognition as a single-label classification task (the weather label means weather category in this paper), namely, determining whether an image belongs to a specific weather category or not. 

\begin{figure}
	\centering
	\subfigure[]{
		\includegraphics[width=3cm, height=1.8cm]{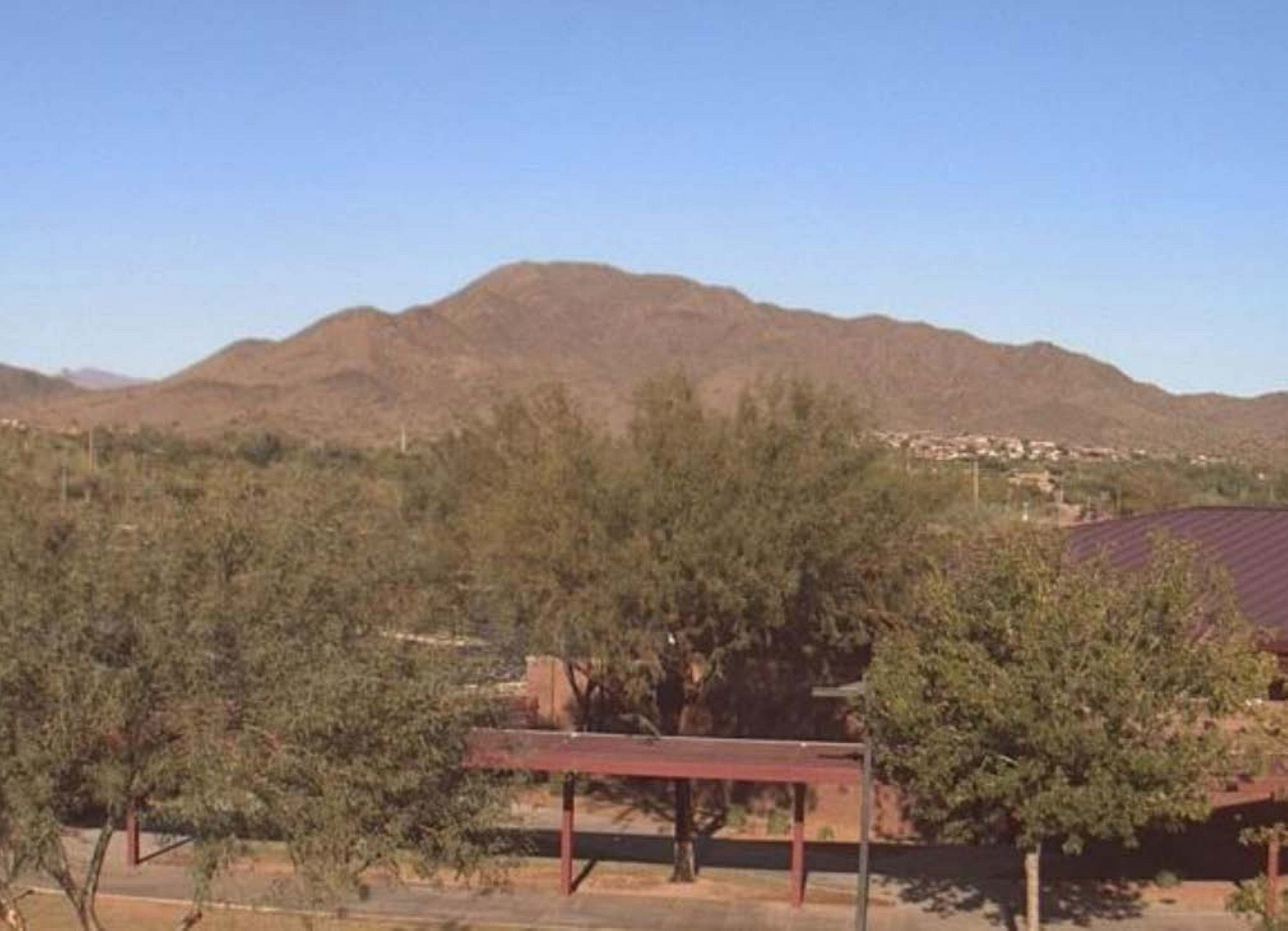}}
	\subfigure[]{
		\includegraphics[width=3cm, height=1.8cm]{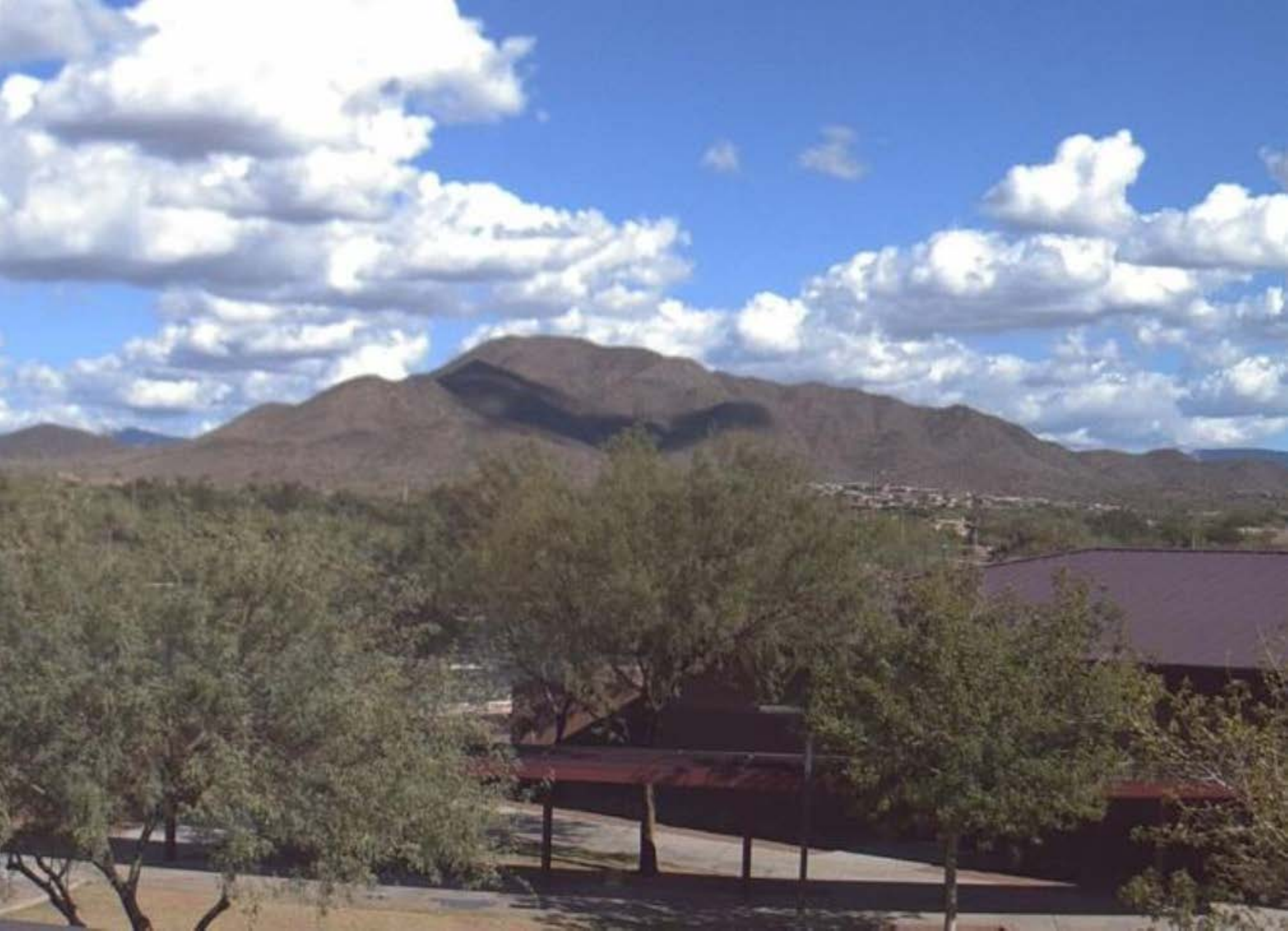}}
	\subfigure[]{
		\includegraphics[width=3cm, height=1.8cm]{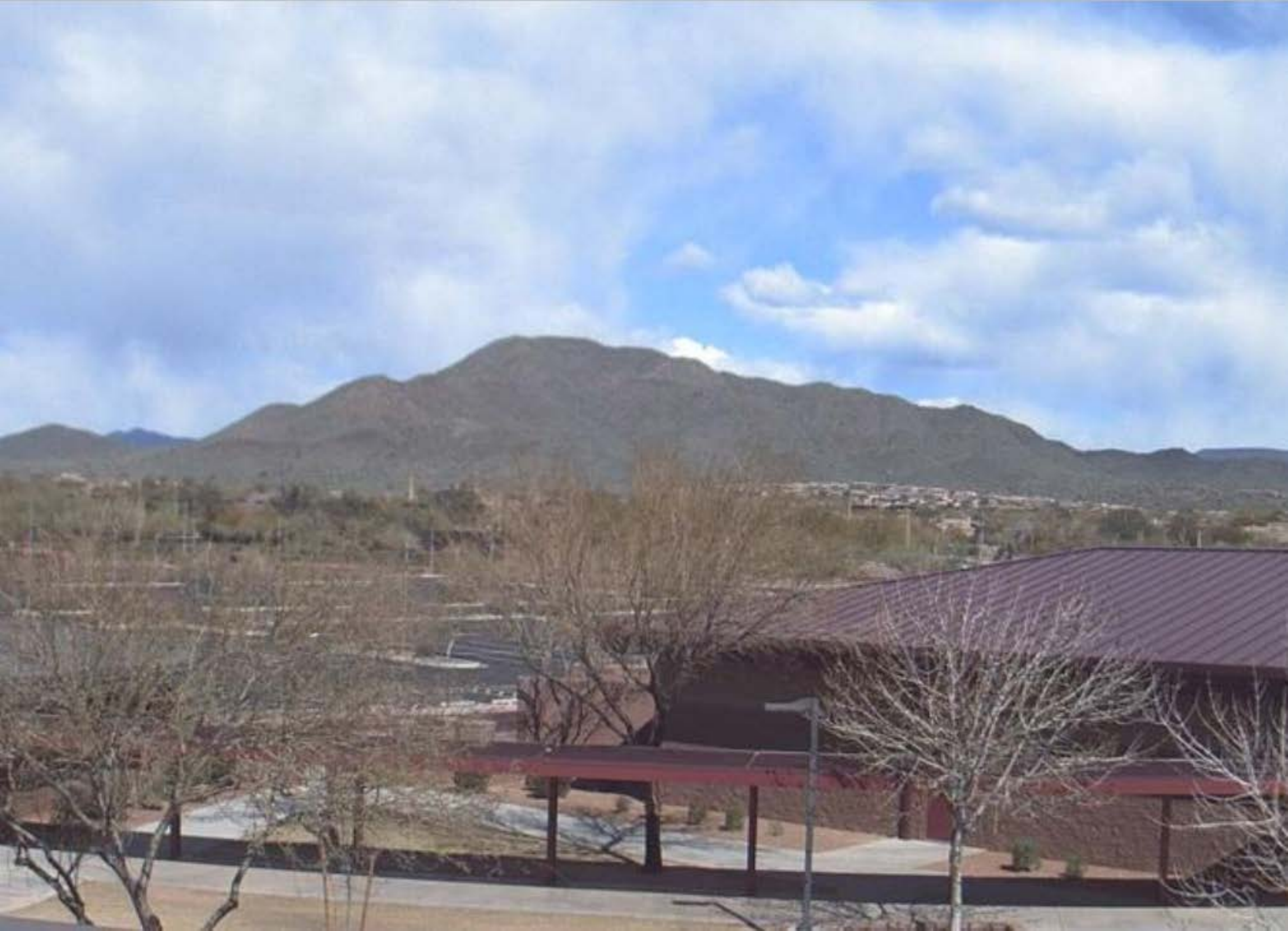}}
	\subfigure[]{
		\includegraphics[width=3cm, height=1.8cm]{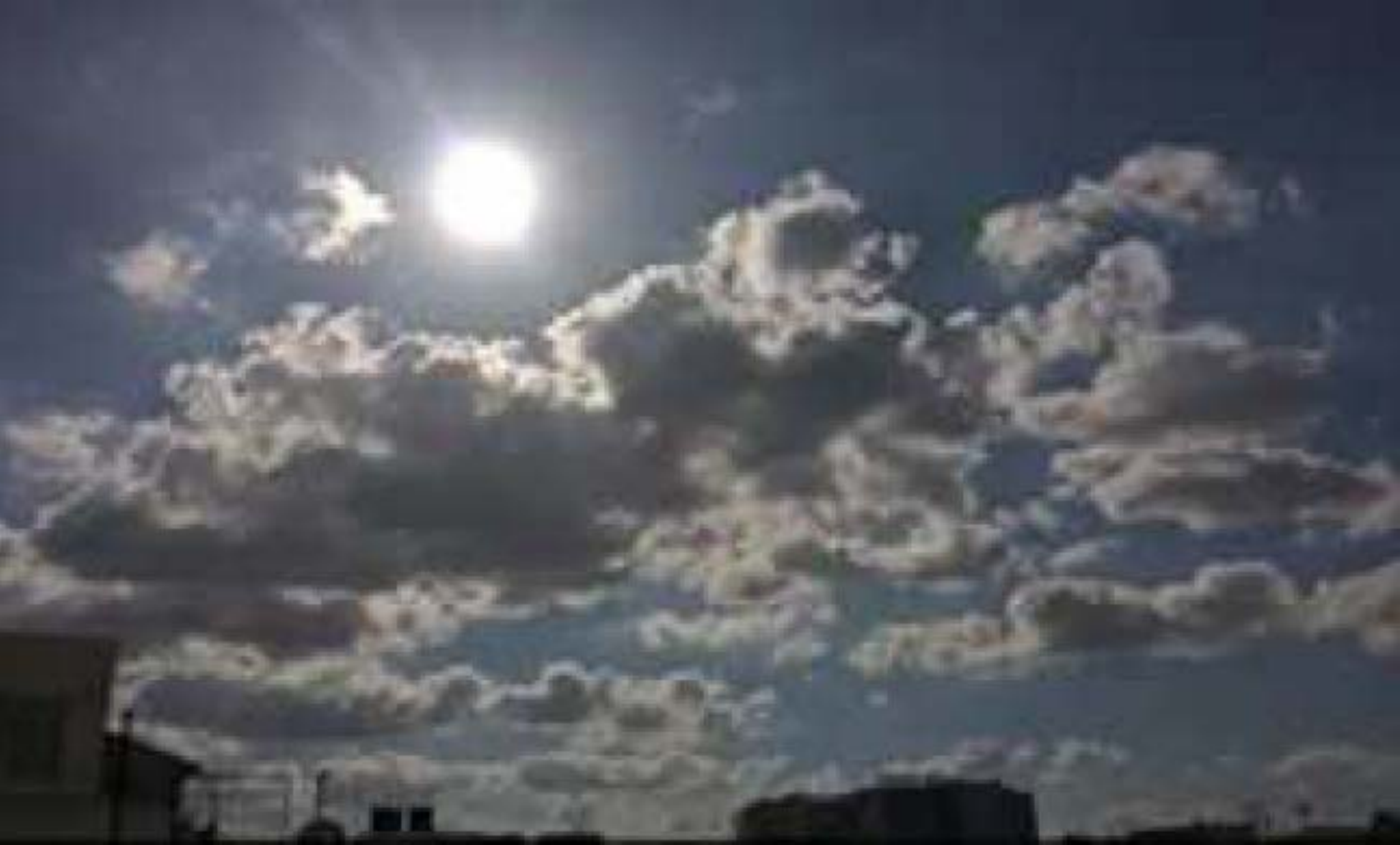}}
	\subfigure[]{
		\includegraphics[width=3cm, height=1.8cm]{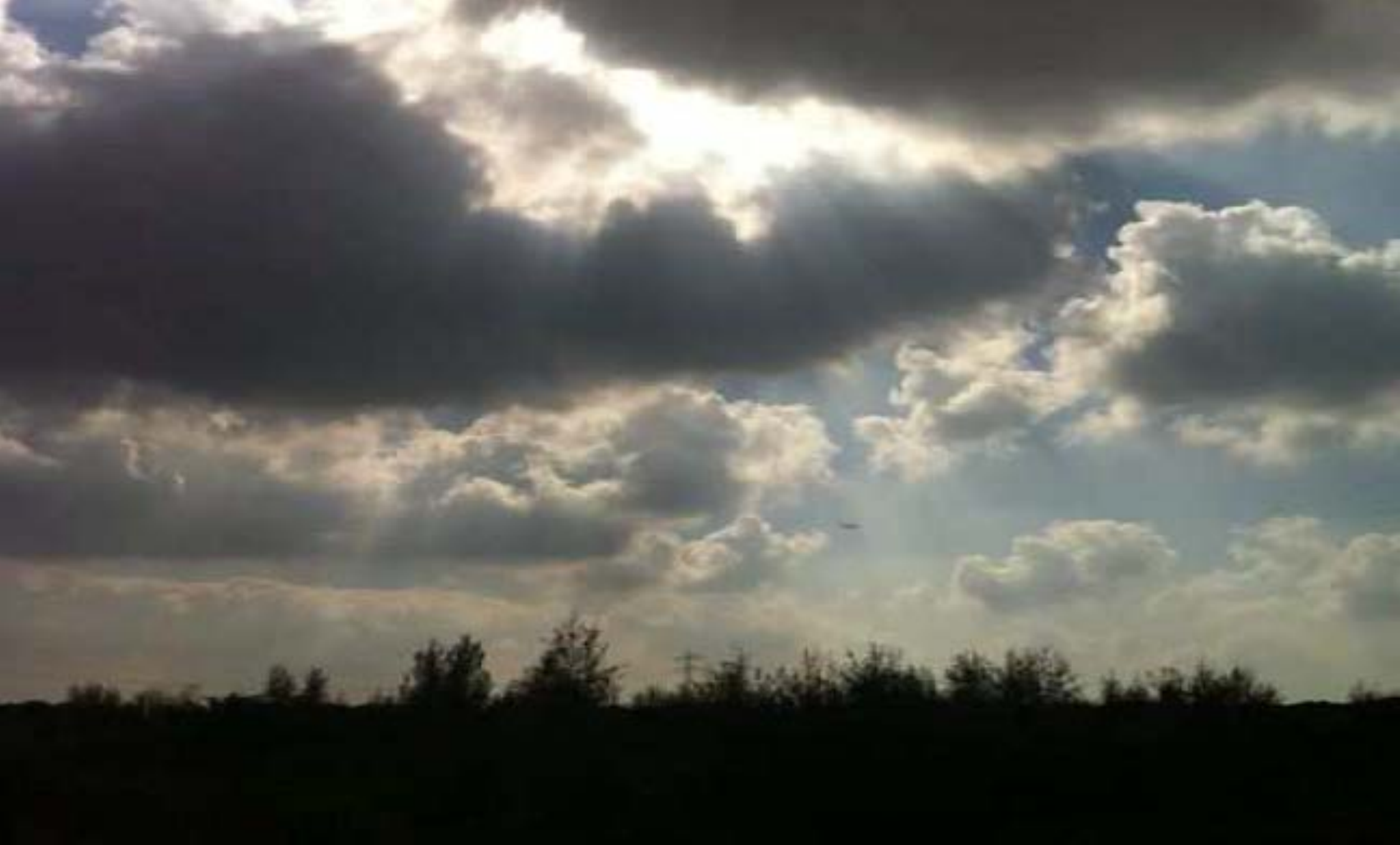}}
	\subfigure[]{
		\includegraphics[width=3cm, height=1.8cm]{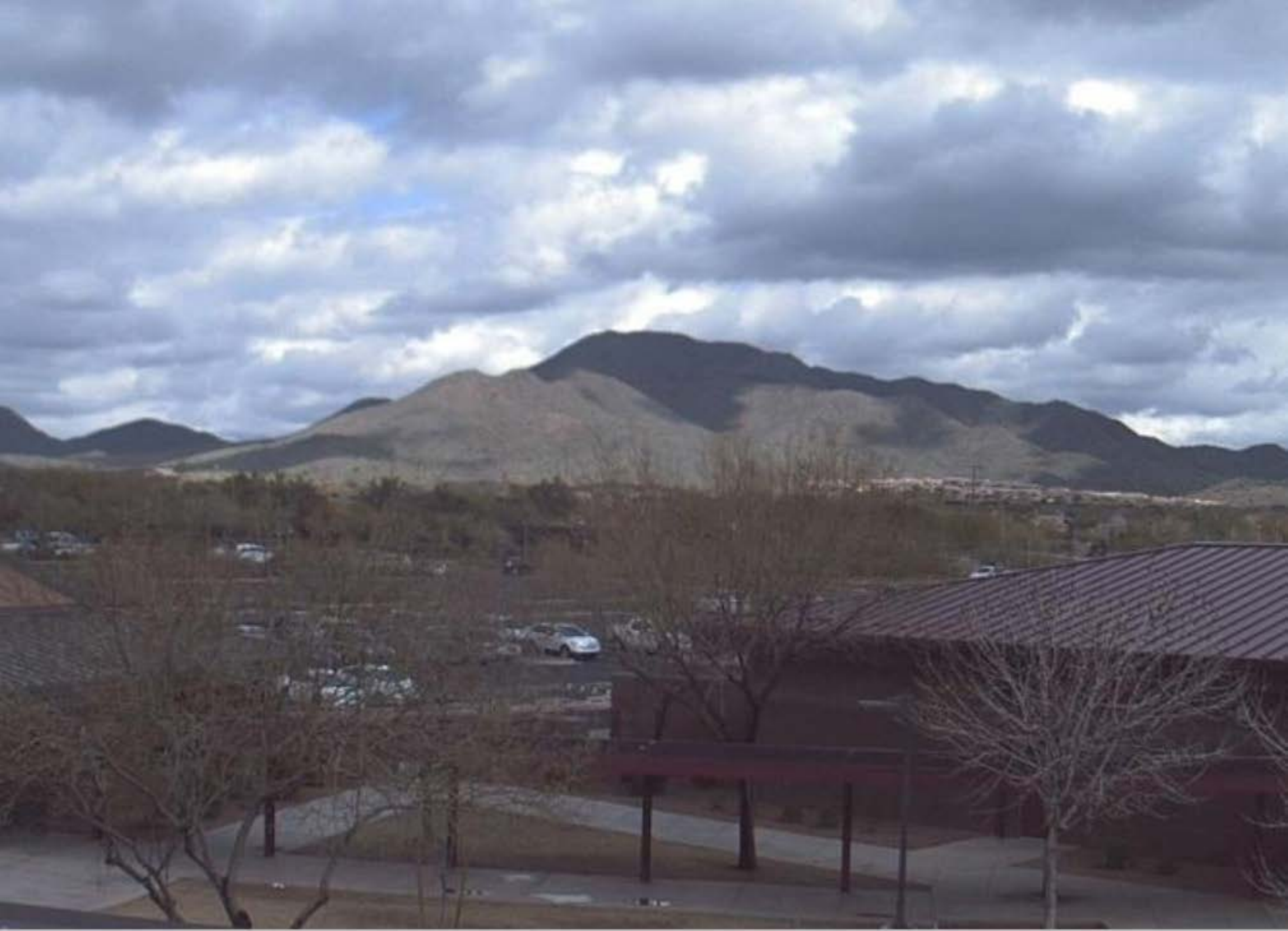}}
	\subfigure[]{
		\includegraphics[width=3cm, height=1.8cm]{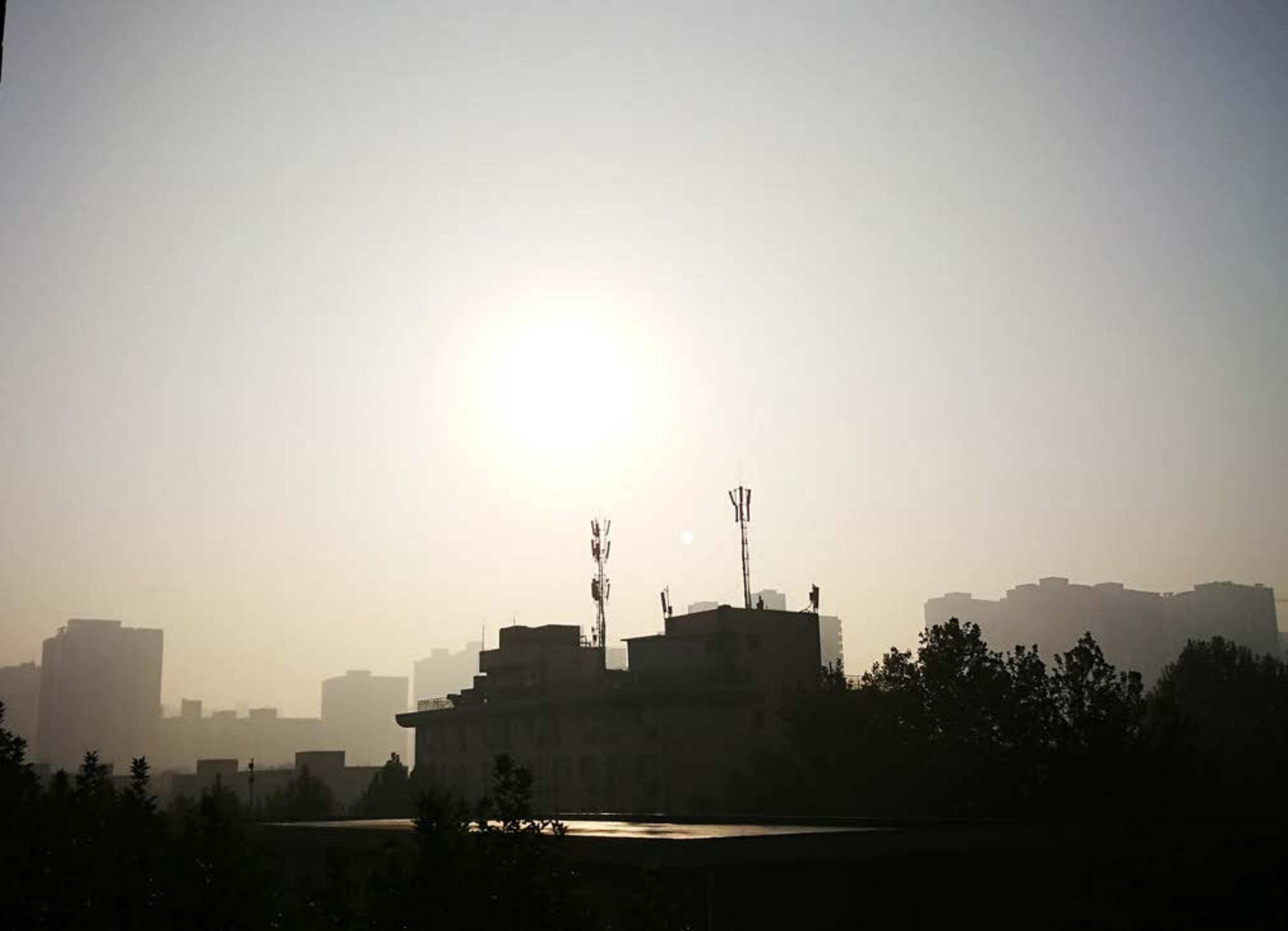}}
	\subfigure[]{
		\includegraphics[width=3cm, height=1.8cm]{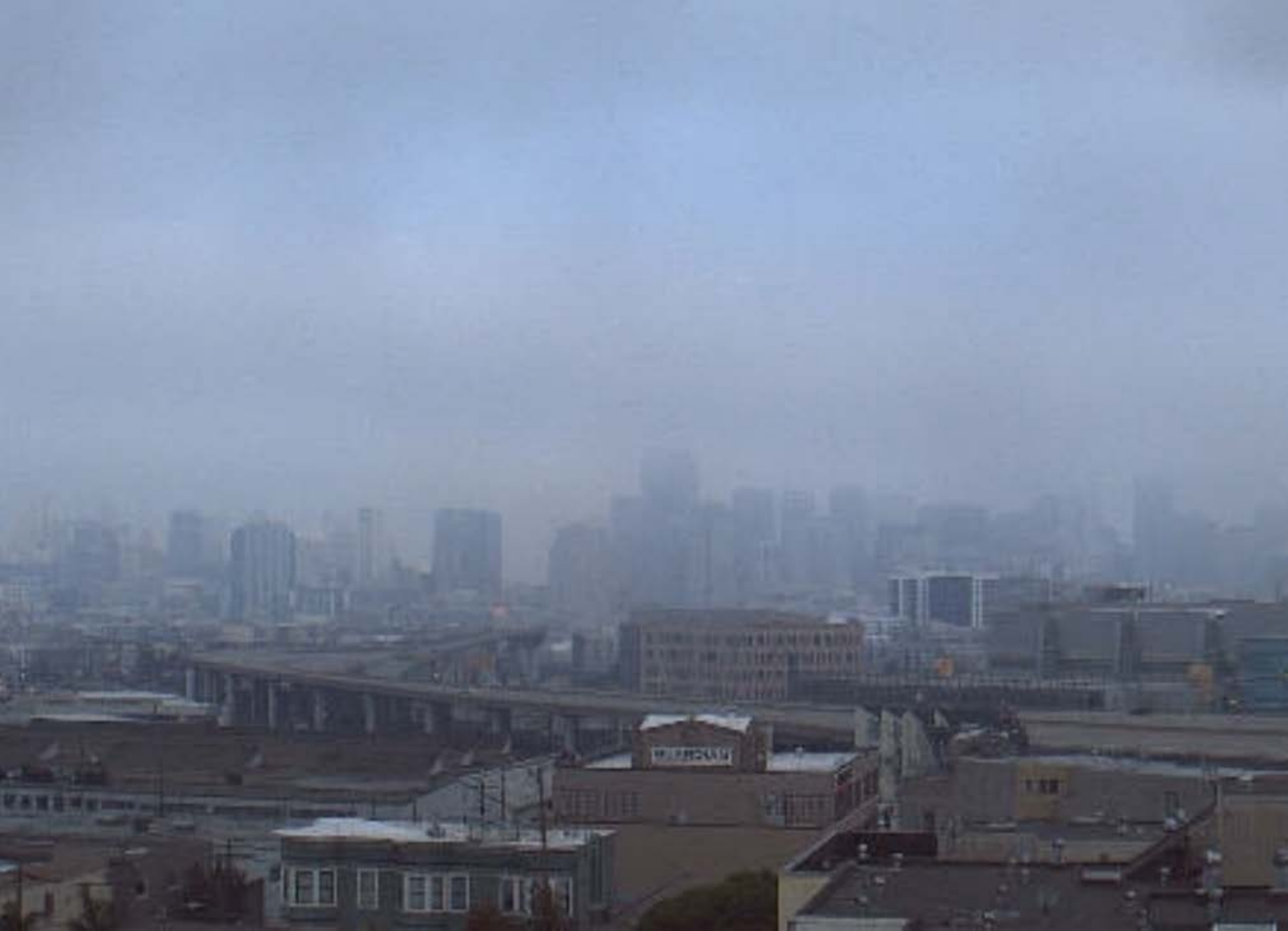}}
	\subfigure[]{
		\includegraphics[width=3cm, height=1.8cm]{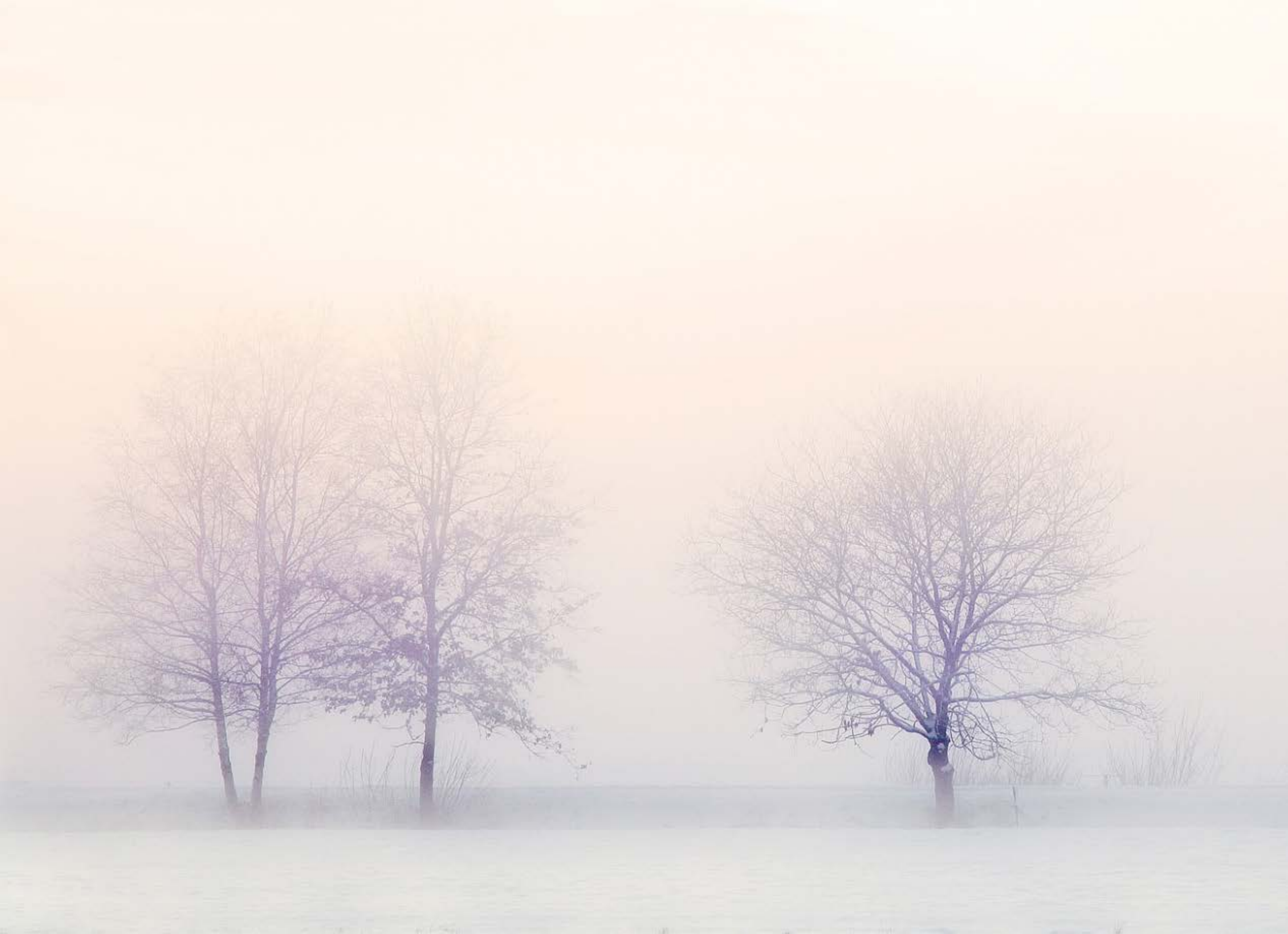}}
	\caption{Several outdoor images (A part of these images are from \cite{Laffont14}). (a) A sunny image with no clouds in the sky. (b) A sunny image with a few clouds in the sky. (c) In comparison with (b), there are more clouds in this image. (d) and (e) present both sunny and cloudy features. (f) A cloudy image. (g) A foggy image with an obvious sun in the sky. (h) A foggy image that seems more overcast. (I) A foggy image with obvious snow on the ground.}
	\label{examples}
\end{figure}

However, it is not always appropriate to view weather recognition as a single-label classification problem. There are mainly two reasons to explain this inappropriateness. The first reason can be summarized as uncertainty, i.e., the class boundaries among some weather categories are ambiguous essentially. As can be seen from Figure \ref{examples}, the changes from Figure \ref{examples} (a) to (f) demonstrate that there are a series of states between a pure sunny weather (like Figure \ref{examples} (a)) and an obvious cloudy weather (as illustrated in Figure \ref{examples} (f)). It is hard to determine whether the category is sunny or cloudy when referring to an intermediate weather state like Figure \ref{examples} (c), (d) and (e) \cite{7784804}. Thus, the uncertainty of such boundary samples causes the difficulty to determine ground-truth labels even from the perspective of human beings, and few previous works present solutions to this problem. The second drawback of treating weather recognition as a single-label classification task can be summarized as incompleteness, namely, a single weather label may not describe the weather conditions comprehensively for a given image. For example, the visual effect of haze is obvious in Figure \ref{examples} (g), (h) and (i). Nevertheless, it can be seen from the comparisons among these three images that Figure \ref{examples} (g) seems more sunny while Figure \ref{examples} (h) seems more overcast, and Figure \ref{examples} (i) seems snowy. Therefore, only a haze label cannot reveal the differences among these three images.

Motivated by the aforementioned two reasons, we propose to view weather recognition as a multi-label classification problem, i.e., assigning multi-labels to an image according to the displayed weather conditions. Specifically, it is achieved by a CNN-RNN architecture. The intuition lies in two aspects. On one hand, most of the previous works focused on exploiting hand-crafted weather features \cite{Lu:2014:TWC:2679600.2680161}, \cite{Zhang2016365}, while these features did not achieve desired results in the weather recognition task. Inspired by the great success of Convolutional Neural Network (CNN) in these years, we utilize CNN as the weather feature extractor. On the other hand, labels exhibit strong co-occurrence dependencies in weather domain. For example, snowy and cloudy usually occur together while rainy and sunny almost never co-occur. Inspired by the success of Recurrent Neural Network (RNN) in dependency modeling \cite{DBLP:conf/aaai/ChenDZH18,DBLP:conf/aaai/ChenDZCLH17}, we propose to use RNN to model the dependencies among labels and predict weather labels step by step. In such a way, when predicting subsequent labels, the network can refer to the previous hidden states that incorporate the historical information implicitly.

\begin{figure}
	\centering
	\subfigure[]{
		\includegraphics[width=4cm]{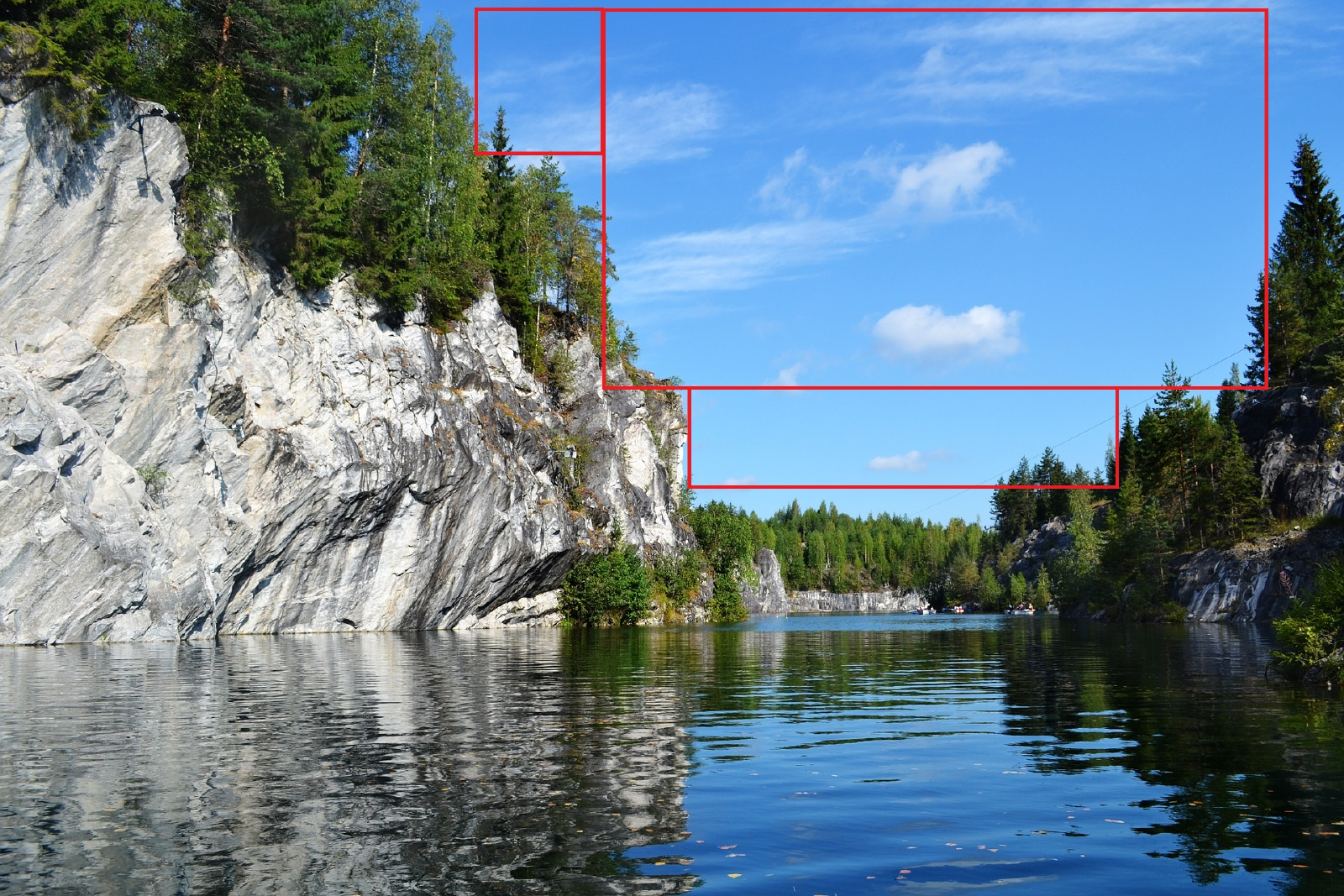}}
	\subfigure[]{
		\includegraphics[width=4cm]{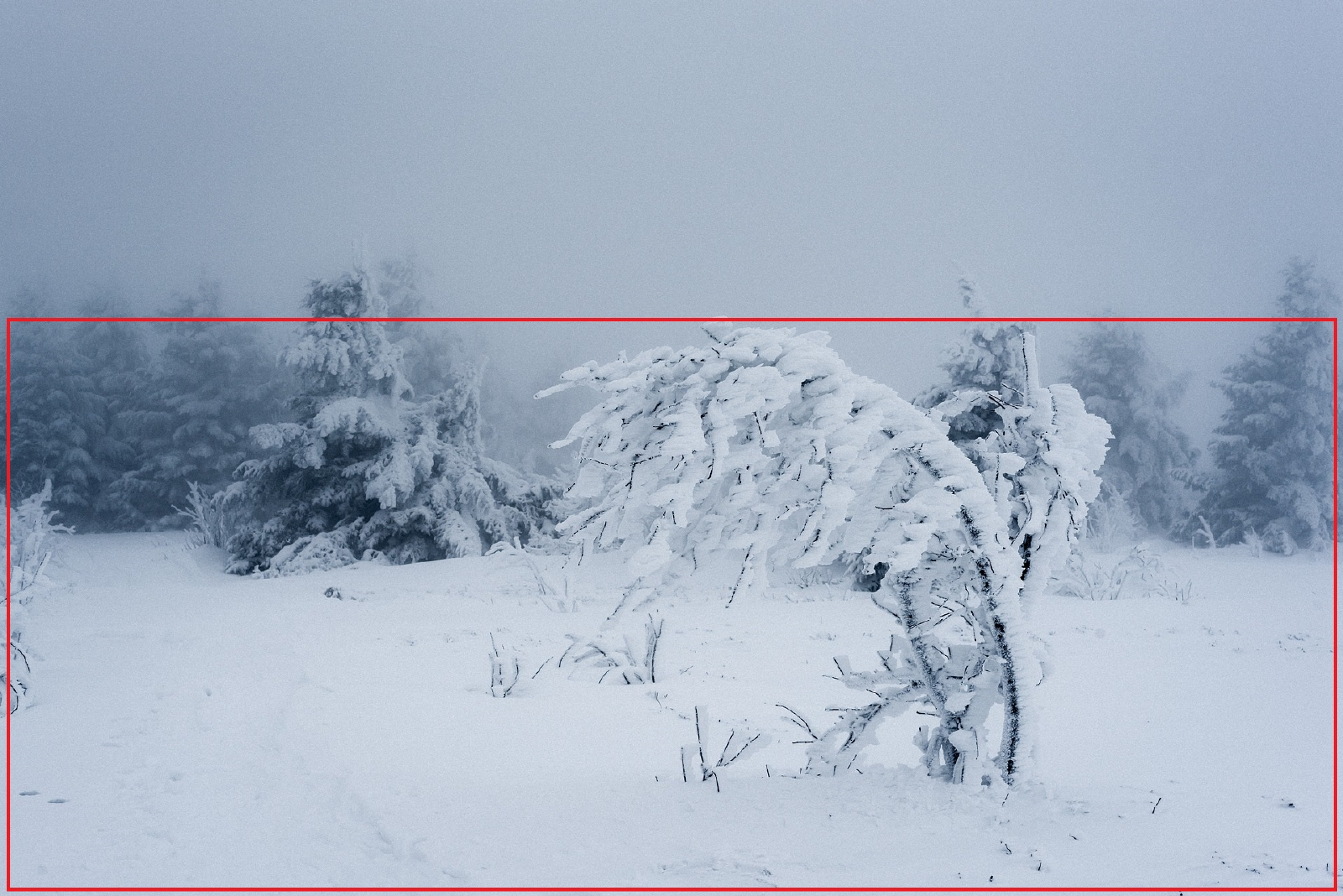}}
	\subfigure[]{
		\includegraphics[width=4cm]{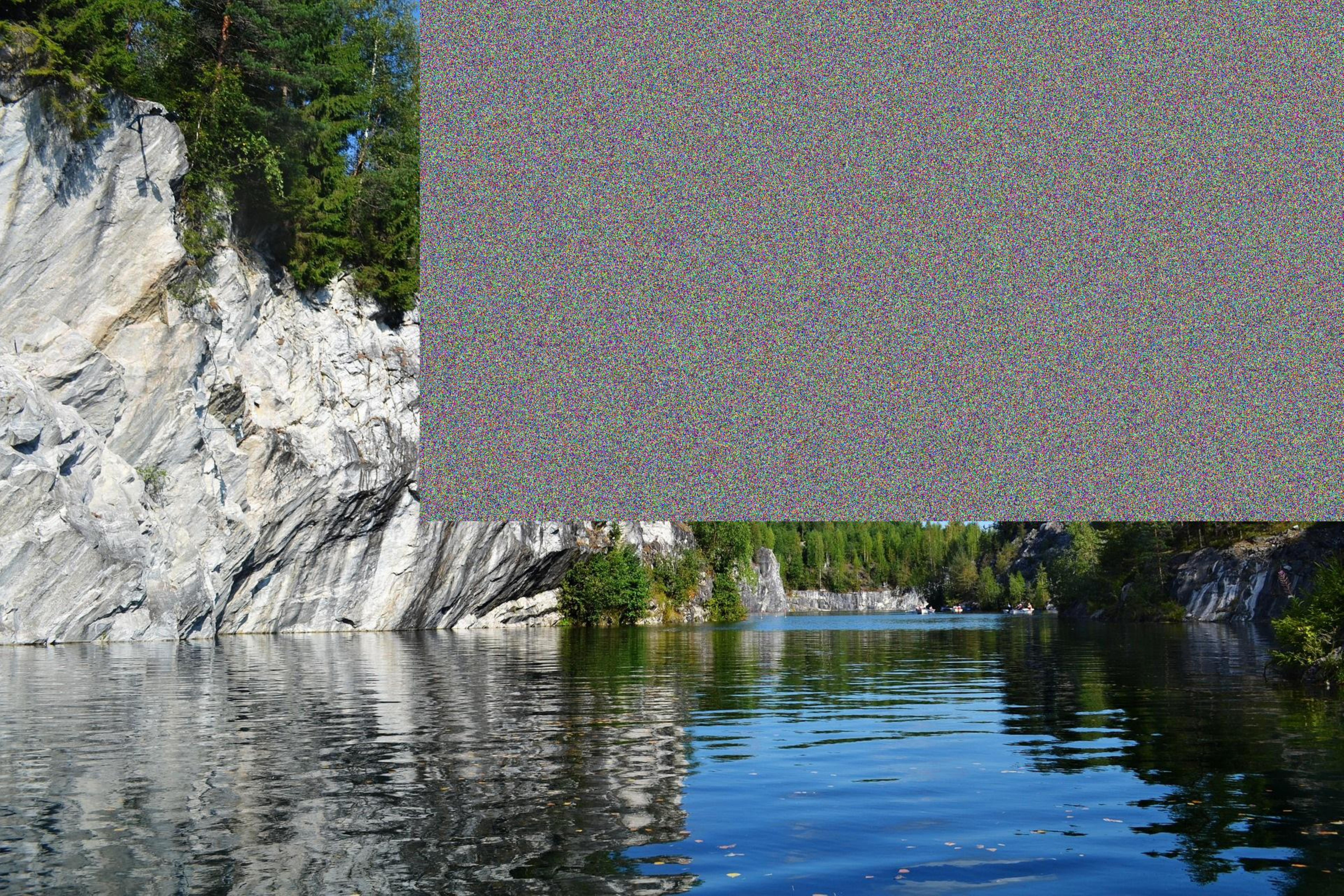}}
	\subfigure[]{
		\includegraphics[width=4cm]{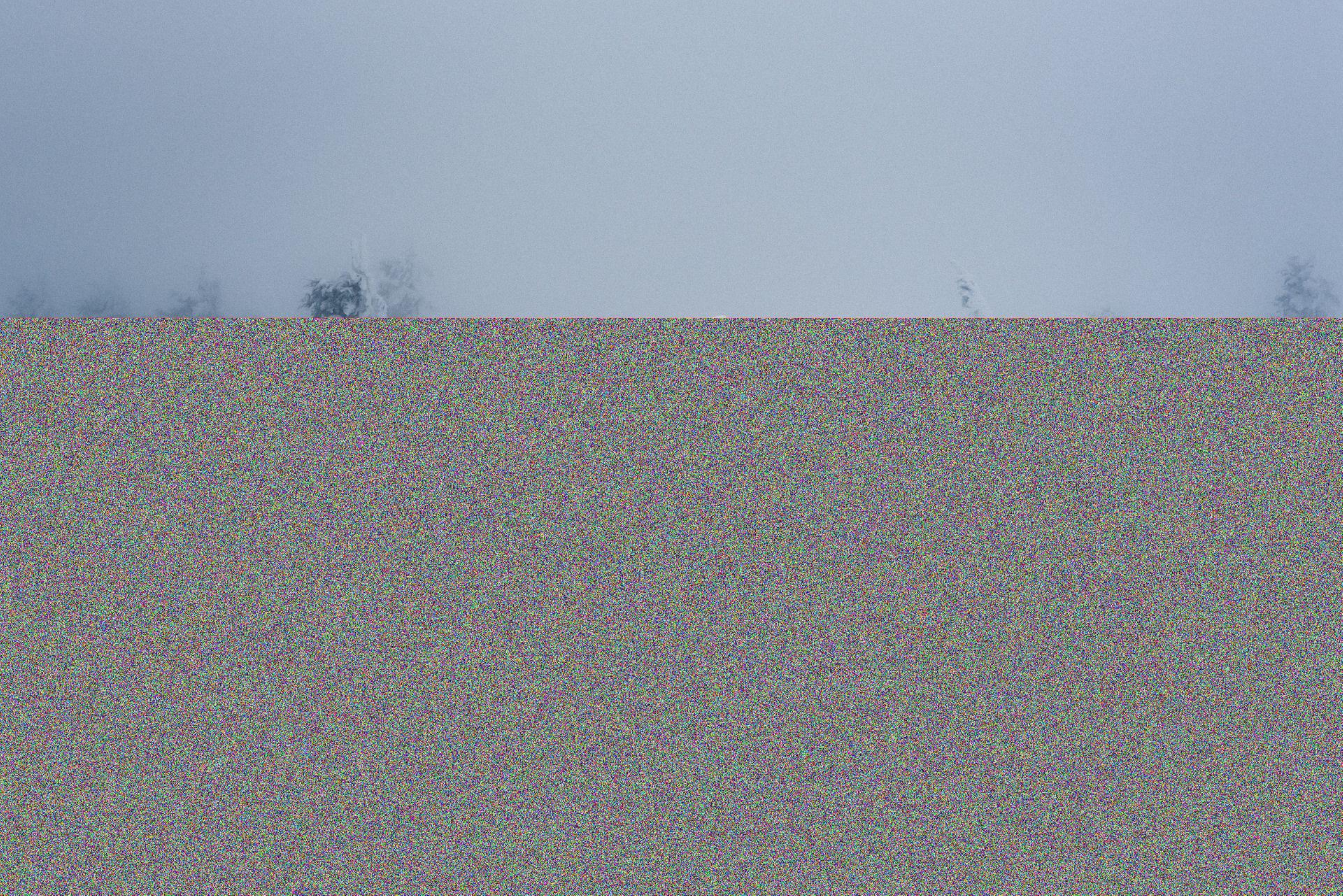}}
	\caption{Illustration of some important regions in weather recognition. (a) A sunny weather with blue sky. (b) A snowy weather with snow on the ground. (c) and (d) When occluding the important regions, it is more difficult to estimate the weather conditions.}
	\label{weather cues}
\end{figure}

For weather recognition, different image regions have different importances when predicting labels. As shown in Figure \ref{weather cues}, the blue sky is crucial for judging a sunny day, and snow on the ground is significant for estimating the snowy weather. Lu \emph{et al.} \cite{7784804} also emphasized that such weather cues are critical. Therefore, it is necessary to make the weather cues discriminative and preserve the spatial information of the image. To achieve this goal, a channel-wise attention model is designed to exploit more discriminative features for the weather recognition task. Besides, we use convolutional Long Short-Term Memory (LSTM) \cite{DBLP:conf/nips/ShiCWYWW15} instead of vanilla RNN in our CNN-RNN architecture to preserve the spatial information. Convolutional LSTM uses convolution operations in both state-to-state and input-to-state transformations, which captures better spatio-temporal information than fully connected LSTM (FC-LSTM) \cite{DBLP:conf/nips/ShiCWYWW15}.

In addition, considering that there are lacking of datasets in the weather recognition task, two new datasets are constructed in this paper, where the first consists of about 8K images from seven weather categories, it is transformed from an existing transient attribute dataset \cite{Laffont14}. The second is built from scratch containing 10K images from five weather categories.

\subsection{Contributions}
In summary, there are three main contributions of this work:
\begin{enumerate}
	\item[(1)]
	We propose to treat weather recognition as a multi-label classification task by analyzing the drawbacks of classifying images with a single weather label and the co-occurrence relationships among different weather conditions.
	\item[(2)] 
	We present a CNN-RNN architecture to tackle the multi-label weather classification task. It is composed of a CNN to extract features, a channel-wise attention model to recalibrate feature responses, and a convolutional LSTM to model the relationships among different weather labels.
	\item[(3)]
	We build a new multi-label weather classification dataset and transform an existing transient attribute dataset \cite{Laffont14} for the weather recognition task. The datasets will be available on the project website. 
\end{enumerate}

\subsection{Organization}
The remainder of this paper is in the following structure: In section \ref{relatedwork} , some related works about weather recognition are reviewed. In section \ref{Approach}, we describe the proposed approach in detail. In section \ref{Experiments}, we first present the construction of the new multi-label weather image dataset and the modification of the transient attribute dataset \cite{Laffont14}. Then, we analyze the experimental results on these two datasets. In section \ref{conclusion}, the conclusion of this paper is drawn.

\section{Related Work}\label{relatedwork}
We roughly classify the weather recognition works into two subcategories in this paper. One category focuses on designing hand-crafted weather features, and another category attempts to use CNNs to solve the weather recognition task.

\subsection{Weather Recognition with hand-crafted features}
Many vehicle assistant driving systems use weather recognition to improve the road safety. For example, they can set speed limit in extreme weather conditions, automatically open the wiper in a rainy day and so forth. Hand-crafted features are popular in these works. Kurihata \emph{et al.} \cite{1505103, 1692045} proposed that rain drops are strong cues for the presence of rainy weather and developed a rain feature to detect rain drops on the windshield. Roser \emph{et al.} \cite{4621205} defined several regions of interest (ROI) and developed various types of histogram features for rainy weather recognition. Yan \emph{et al.} \cite{Yan:2009:WRB:1561386.1561431} utilized gradient amplitude histogram, HSV color histogram as well as road information for the classification task among sunny, cloudy and rainy categories. Besides, several methods are proposed specially for fog detection, Hauti{\'e}re \emph{et al.} \cite{Hautiére2006} used Koschmieder’s Law \cite{Middleton1957} to detect the presence of fog and estimate the visibility distance. Bronte \emph{et al.} \cite{5309842} utilized many techniques, including a Sobel based sunny-foggy detector, edge binarization, hough line detection, vanishing point detection and road/sky segmentation. Gallen \emph{et al.} \cite{5940486} focused on night fog detection by detecting backscattered veil caused by the vehicle ego lights or halos around the street lights. Pavlić \emph{et al.} \cite{6232256, DBLP:conf/ivs/PavlicRI13} transformed images into frequency domain and detected the presence of fog through training different scaled and oriented Gabor filters in the power spectrum. Although the aforementioned approaches have shown good performance, they are usually limited to the in-vehicle perspective and cannot be applied to  wider range of applications. 

There are also several researches devoted to weather recognition from common outdoor images. Li \emph{et al.} \cite{5206732} proposed a photometric stereo-based approach to estimate weather condition of a given site. Zhao \emph{et al.} \cite{6115972} pointed out that pixel-wise intensities of dynamic weather conditions (rainy, snowy, etc.) fluctuate over time while static weather conditions (sunny, foggy, etc.) almost stay unchanged. They proposed a two-stage classification scheme which first distinguishes between the two conditions then utilizes several spatio-temporal and chromatic features to further estimate the weather category. In \cite{Song2014}, several global features were extracted for weather classification, such as inflection point information, power spectral slope, edge gradient energy, saturation, contrast and image noise. Li \emph{et al.} \cite{7294972} also utilized several features in \cite{Song2014}, and constructed a decision tree according to the distance between features. Except for regular global features, \cite{Lu:2014:TWC:2679600.2680161} proposed multiple weather cues including reflection, shadow and sky descriptor for two-class weather recognition. They also exploited a collaborative learning strategy in which voters closer to the test image have more weights. Zhang \emph{et al.} \cite{7351637, Zhang2016365} proposed the sunny feature, rainy feature, snowy feature and haze feature individually for each weather class as well as two global features. Furthermore, a multiple kernel learning approach is proposed in \cite{7351637} to fuse these features. In \cite{6247815}, both spatial appearance and temporal dynamics were investigated on short video clips which can recognize several weather types.

Although researchers have elaborately designed many features for weather recognition, the features are usually limited to specific perspectives or weather classes, and cannot be applied to wider range of applications.

\subsection{Weather Recognition with CNNs}
In recent years, convolutional neural networks have shown overwhelming performance in a variety of computer vision tasks, such as image classification \cite{NIPS2012_4824}, object detection \cite{renNIPS15fasterrcnn}, semantic segmentation \cite{8237584}, etc. Several excellent architectures of CNNs are proposed including AlexNet \cite{NIPS2012_4824}, VGGNet \cite{Simonyan14c} and ResNet \cite{7780459}, which outperform the traditional approaches to a large extent. Inspired by the great success of CNNs, a few of works attempt to apply CNNs to weather recognition task. Elhoseiny \emph{et al.} \cite{7351424} directly fine-tuned AlexNet \cite{NIPS2012_4824}  on a two-class weather classification dataset released by \cite{Lu:2014:TWC:2679600.2680161}, and achieved a better result. Lu \emph{et al.} \cite{7784804} combined hand-crafted weather features with CNNs extracted features, and further improved the classification performance. While as discussed in \cite{7784804}, there is no closed boundaries among weather classes. Multiple weather conditions may appear simultaneously. Therefore, all the above approaches suffer from the information loss when they treat weather recognition as a single label classification problem. Li \emph{et al.} \cite{DBLP:conf/mm/LiWL17} proposed to use auxiliary semantic segmentation of weather cues to comprehensively describe the weather conditions. This strategy can alleviate the problem of information loss, while the segmentation mask is not intuitive for humans. 

\section{Our Approach}\label{Approach}

\begin{figure*}
	\centering
	\includegraphics[width=1\textwidth]{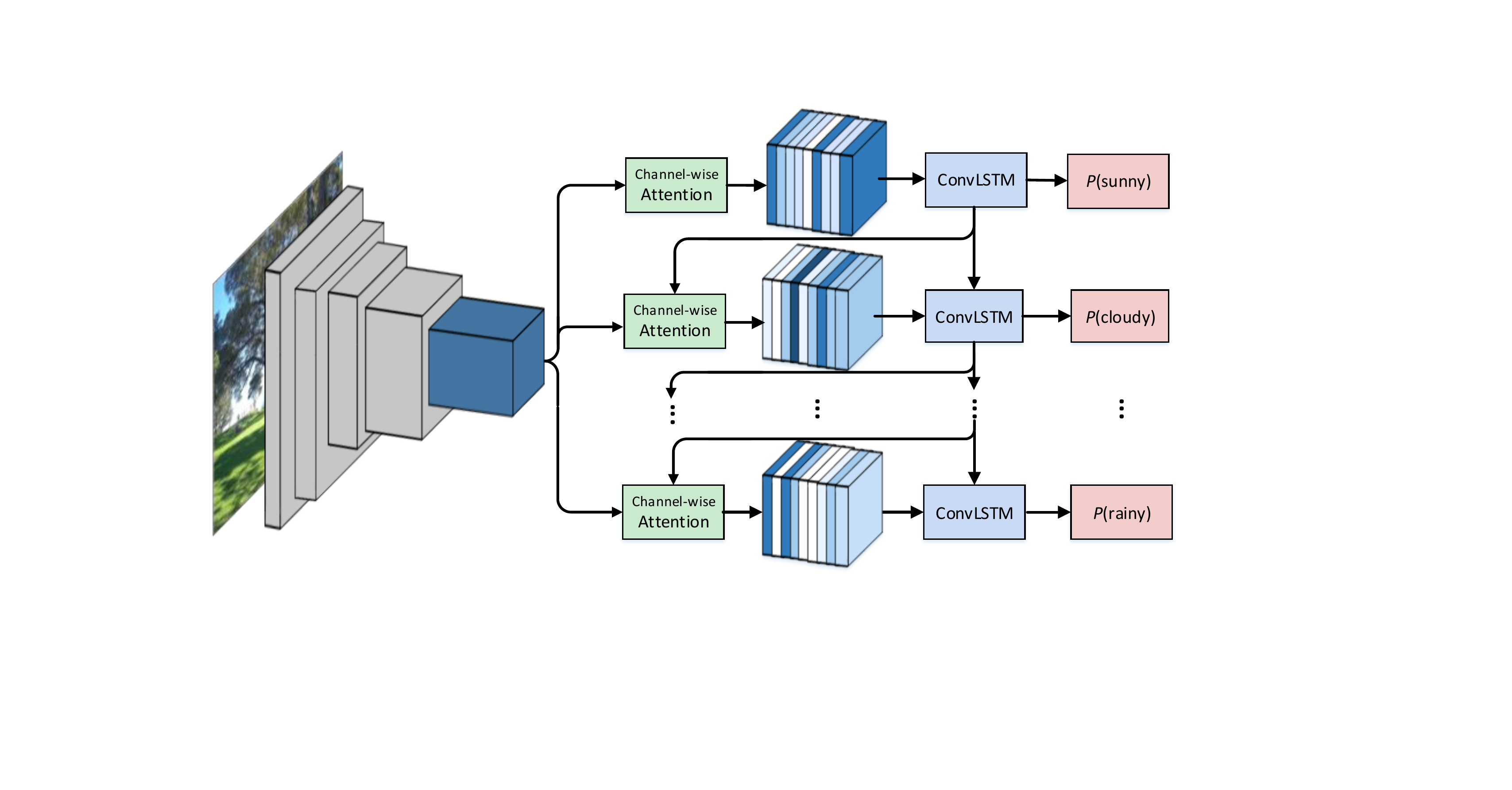}
	\caption{The illustration of the proposed CNN-RNN architecture for multi-label weather classification. First, the CNN is employed to extract image features. Then, the convolutional LSTM is used to predict weather labels step by step. In each step, a channel-wise attention model is utilized to recalibrate the feature responses. Each output label is estimated according to the adjusted features and the hidden state which implicitly maintains the information of previously predicted labels.}
	\label{architecture}
\end{figure*}

In this paper, to comprehensively describe the weather conditions, we propose to treat weather recognition as a multi-label classification problem. Furthermore, a CNN-RNN model is developed for this task, which formulates the multi-label classification as a step-wise prediction. Figure \ref{architecture} demonstrates the architecture of the proposed approach. It mainly composes of three parts, i.e., the basic CNN, a channel-wise attention model and a convolutional LSTM. The CNN extracts the preliminary features of a given outdoor image. Specifically, the first five groups of convolutional/pooling layers of VGGNet \cite{Simonyan14c} are utilized in this paper. The channel-wise attention model adaptively calculates the channel-wise attention weights and recalibrates the feature responses. The convolutional LSTM uses visual features and the hidden state to predict weather labels one by one, which implicitly models the co-occurrence dependency among labels by maintaining context information in internal memory states.

\subsection{The Convolutional LSTM in the CNN-RNN Architecture}
The Recurrent Neural Networks, especially LSTM, has recently achieved overwhelming success in sequence modeling tasks, such as image/video captioning \cite{ijcai2017-307} and neural machine translation \cite{DBLP:journals/corr/BritzGLL17}. Without loss of generality, the LSTM can be formulated as follows \cite{hochreiter1997long}.
\begin{align}
	\label{lstm}
	\begin{split}
		&i_t=\sigma(W_{iw}x_t+U_{ih}h_{t-1}+b_i),
		\\
		&f_t=\sigma(W_{fw}x_t+U_{fh}h_{t-1}+b_f),
		\\
		&o_t=\sigma(W_{ow}x_t+U_{oh}h_{t-1}+b_o),
		\\
		&g_t=\tanh(W_{gw}x_t+U_{gh}h_{t-1}+b_g),
		\\
		&c_t=f_t\circ c_{t-1}+i_t\circ g_t,
		\\
		&h_t=o_t\circ \tanh c_t,
	\end{split}
\end{align}
where the subscript $t$ indicates the $t$-th step of LSTM, $x_t$ denotes the input data, $h_t$ stands for the hidden state, $c_t$ is the cell state. $i_t$, $f_t$ and $o_t$ are input gate, forget gate and output gate of the LSTM, respectively. $W$s , $U$s and $b$s are weights and biases to be learned. $\sigma$, $\tanh$ and $\circ$ represent the sigmoid function, hyperbolic tangent function and element-wise multiplication, respectively. As shown in Eq. 1, at each step, the data \(x_t\) and the previous hidden state \(h_{t-1}\) is taken as the input of current LSTM unit, and the historical information are recorded in the hidden state \(h_t\), such that LSTM can exploit the temporal dependency.

Although the standard LSTM has demonstrated its powerful capability in sequence modeling tasks, the spatial information is ignored when processing images \cite{DBLP:conf/nips/ShiCWYWW15}. As can be seen from Eq. \ref{lstm}, fully connections are used in state-to-state and input-to-state transformations. Generally, if the input image data $x_t \in R^{W\times H \times C}$, it will be flattened to an 1D-vector before input to the LSTM. While this process will suffer from the loss of spatial information. To overcome this drawback, the convolutional LSTM is employed in our approach \cite{DBLP:conf/nips/ShiCWYWW15}, which can be formulated as follows.
\begin{align}
	\label{convlstm}
	\begin{split}
		&i_t=\sigma(W_{iw} \otimes x_t+U_{ih} \otimes h_{t-1}+b_i),
		\\
		&f_t=\sigma(W_{fw} \otimes x_t+U_{fh} \otimes h_{t-1}+b_f),
		\\
		&o_t=\sigma(W_{ow} \otimes x_t+U_{oh} \otimes h_{t-1}+b_o),
		\\
		&g_t=\tanh(W_{gw} \otimes x_t+U_{gh} \otimes h_{t-1}+b_g),
		\\
		&c_t=f_t\circ c_{t-1}+i_t\circ g_t,
		\\
		&h_t=o_t\circ \tanh c_t,
	\end{split}
\end{align}
where $\otimes$ denotes the convolution operator and other symbols are the same with Eq. \ref{lstm}. It should be noted that the input feature $x_t$, cell state $c_t$, hidden state $h_t$ and gates $i_t$, $f_t$, $o_t$ of convolutional LSTM are all 3D tensors, and convolution operations are used in state-to-state and input-to-state transformations. Therefore, the spatial information of features are preserved in this way. Furthermore, the convolution operation actually has implicit spatial attention mechanism, since regions corresponding to the target label usually have higher activation responses. In the experiment, we also find that the convolutional LSTM pays attention to several critical regions for weather label prediction, and achieves better results than common LSTM with or without spatial attention model. 

\subsection{Channel-wise Attention Model in the CNN-RNN Architecture}
\begin{figure}
	\centering
	\includegraphics[width=9.5cm]{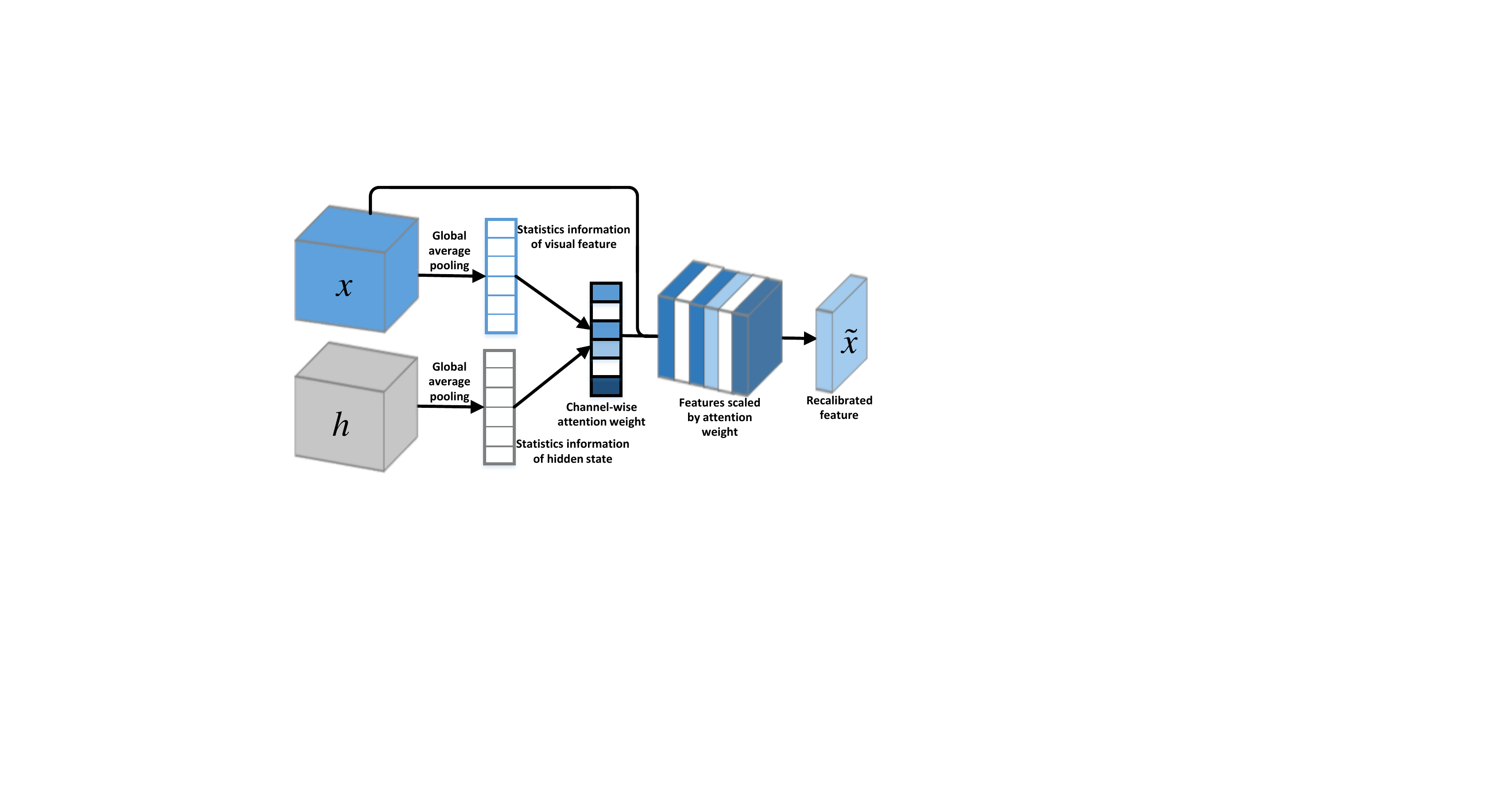}
	\caption{The illustration of the proposed channel-wise attention model for CNN-RNN architecture. It takes the visual feature \(x\) and the hidden state \(h\) as input, and outputs the recalibrated feature \(\tilde x\).}
	\label{attention}
\end{figure}

Usually, different regions will be activated in disparate channels of the feature map, and different image regions have different importance when estimating various weather conditions. In our CNN-RNN architecture, each step of the convolutional LSTM will predict one weather label. Inspired by \cite{hu2017}, we propose a channel-wise attention model for the CNN-RNN architecture to adaptively recalibrate the feature responses when predicting different weather labels. The illustration of the proposed channel-wise attention model is shown in Figure \ref{attention}.

As discussed in \cite{hu2017}, exploiting global information is a popular method in feature engineering works. To calculate the attention weight of each feature map channel, we adopt the similar strategy, i.e., global average pooling is used to generate channel-wise statistics which can be viewed as a descriptor of the channel-wise global spatial information. While different from \cite{hu2017}, in our multi-label weather classification task, we want to adaptively obtain the channel-wise attention weights according to the previous predicted weather label. So we also take into account the channel-wise statistics information encoded in the hidden state of the convolutional LSTM. The two kinds of statistics information are formulated as follows.

\begin{equation}
a_k=f_a(x_k)=\frac{1}{W\times H}\sum_{i=1}^W\sum_{j=1}^H x_k(i,j),
\end{equation}
\begin{equation}
d_k=f_a(h_{t-1,k})=\frac{1}{W\times H}\sum_{i=1}^W\sum_{j=1}^H h_{t-1,k}(i,j),
\end{equation}
where $x_k$ and $h_{t-1,k}$ denote the visual feature and the previous hidden state of the convolutional LSTM at the \(k\)-th channel (\(k=1,2,...,C\)), respectively. $f_a$ represents the global average pooling function, $a_k$ and $d_k$ denote the statistics information of visual feature and hidden state at the \(k\)-th channel. $W$ and $H$ stand for the width and height of visual features. It should be noted that, in our approach, the visual features and hidden states are in the same dimension.

After the statistics information of the visual features and hidden states is obtained, the channel-wise attention weights are calculated by 
 \begin{equation}{z_k = \sigma ({w_2}\delta ({w_1}[{a_k},{d_k}] + {b_1}) + {b_2}),}\end{equation}
where $w$s and $b$s are weights and biases to be learned, $\delta$ represents the ReLU \cite{DBLP:conf/icml/NairH10} function that is utilized to learn the non-linear mapping, $[\cdot,\cdot]$ is the concatenation operation, $\sigma$ indicates the sigmoid function which normalizes the attention weight to the range of 0$\sim$1. Finally, the recalibrated features are obtained by rescaling the original features with attention weights,
\begin{equation}
\tilde x = \sum\limits_{k = 1}^C {z_k{x_k}}.
\end{equation}

\subsection{Inference}
In this paper, the weather labels are predicted in a fixed path. Practically, the order of other weather labels are set according to their co-occurrence relationships, details are depicted in Section \ref{Co-occurrence Relationships}.

In each step of the convolutional LSTM, the 3D hidden state is flattened to a 1D vector, then it is used to predict the weather label.
\begin{equation}
p_t=\sigma(w_ph_t+b_p),
\end{equation}
where $p_t \in [0,1]$ is the output probability of the $t$-th weather label, $h_t$ is the flattened hidden state, $w_p$ and $b_p$ are the learned weight and bias.

The loss of each prediction step is determined by the following function.
\begin{equation}
loss_t=\frac{-1}{N}\sum_{i=1}^Np_{i,t}\log\widetilde{p}_{i,t}+(1-p_{i,t})\log(1-\widetilde{p}_{i,t}),
\end{equation}
where $N$ denotes the number of training samples, $p_{i,t}$ indicates the ground-truth label of the $i$-th sample on the $t$-th weather class, and $\widetilde{p}_{i,t}$ is the corresponding predicted label. Finally, the total loss is formulated as follows,
\begin{equation}
Loss=\sum_{t=1}^Tloss_t,
\end{equation}
where $T$ represents the number of all weather classes.

\subsection{Training Details}

The open source library tensorflow is used to implement the proposed approach. To accelerate the convergence, we adopt a two stage training strategy. In the first stage, the basic CNN of our approach (i.e., the first five groups of convolutional/pooling layers of VGGNet \cite{Simonyan14c}) is trained. Specifically, we transform VGGNet into a multi-label classification framework by replacing the output layer with $T$ neurons ($T $ represents the number of weather classes), and train it with multi-label sigmoid cross-entropy loss function. The pre-trained VGGNet model on ImageNet Large Scale Visual Recognition Challenge (ILSVRC) is used for fine-tune. In the second stage, we remove the fully connection layers of VGGNet, and fix the other parameters. Then, the convolutional LSTM and channel-wise attention model are trained from scratch based on the CNN extracted features. Xavier initialization method is employed in this stage. Adam \cite{DBLP:journals/corr/KingmaB14} optimization approach is used to minimize loss functions in both two stages where the first and second momentum are set to 0.9 and 0.999, respectively. To avoid overfitting, the dropout \cite{JMLR:v15:srivastava14a} operation is used after the fully connection layers in both two stages, and $L_2$ regularization is also employed for all weight parameters. We set the dropout ratio and weight of $L_2$ regularization to 0.5 and 0.0005 during the entire training process. The learning rate is initialized as 0.0001 and drops by a factor of 10 after the loss is stable. Besides, we also attempt to fine-tune all parameters after the second training stage, i.e., unfix the parameters of the basic CNN, while experiments prove that this strategy cannot bring performance improvements. 

Before training, each sample is resized into a 256 $\times$ 256 image. Random flip, random crop and random noise are used for data augmentation. We adopt the stochastic mini-batch training strategy, images are randomly shuffled and they constitute mini-batches of size 50 before each training epoch. Table \ref{detail_table} shows the detailed shapes of several critical components of the proposed CNN-RNN architecture. Besides, the shapes of all biases can be easily inferred.

\begin{table*}
	\centering
	\caption{The details of the proposed CNN-RNN architecture.}
	\resizebox{\linewidth}{!}{\begin{tabular}{|c|c|c|c|c|c|}  
			\hline
			Name   & channel-wise attention model & ConvLSTM  &convolution kernel &$w_1$  &$w_2$ \\ \hline
			Shape &$14 \times 14 \times 512$  &$14 \times 14 \times 512$ &$3 \times 3$ &$1024 \times 512$ &$512 \times 512$ \\ \hline
		\end{tabular}}
		\label{detail_table}
	\end{table*}

\section{Experiments}\label{Experiments}
Since this is the first work to treat weather recognition as a multi-label classification problem, there are no existing datasets for this task. Therefore, to evaluate the proposed approach, we construct two datasets where one is the modification of the transient attribute dataset \cite{Laffont14} and another one is created from scratch. In this section, we first introduce the construction procedure and details of the two datasets. Then, the co-occurrence relationships among weather labels are explored. Finally, the evaluation metrics, comparison approaches and experimental results are presented in turn. 

\subsection{Dataset Description}
\subsubsection{The Transient Attribute Dataset}
\begin{figure*}
		\centering
		\includegraphics[width=1\textwidth]{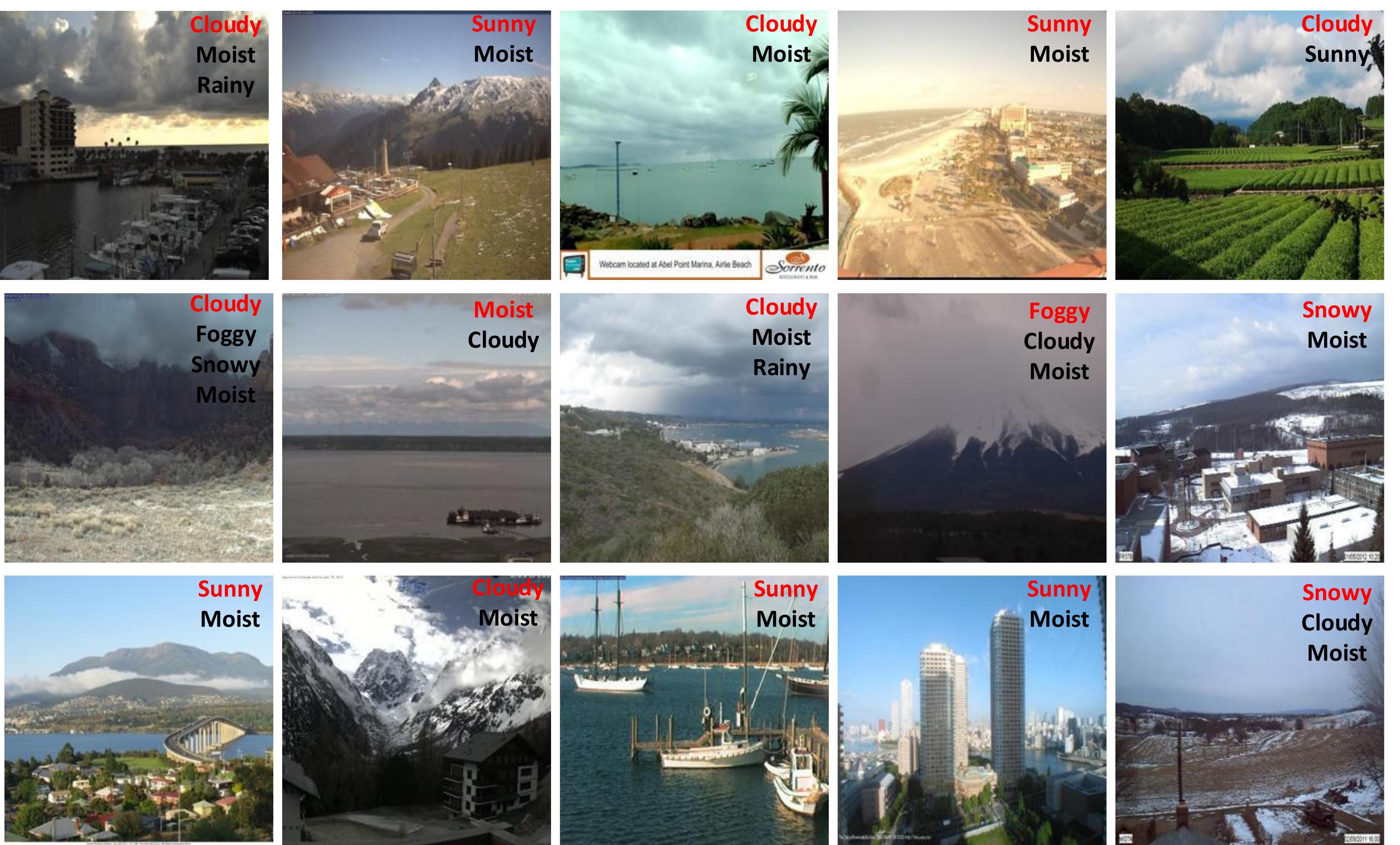}
		\caption{The illustration of examples in the transient attribute dataset. The upper-right of each image displays the studied weather labels. Note that the red label denotes the weather class with the maximum strength, which is usually taken as a single weather label in traditional approaches. Obviously, the multiple weather labels can describe the weather conditions more comprehensively.}
		\label{transient_example}
	\end{figure*}

The first dataset is transformed from an existing transient attribute dataset \cite{Laffont14} which was originally erected for outdoor scenes understanding and editing. Although the transient attribute dataset is not specially designed for weather recognition, this dataset presents many appealing properties. First, images are captured across many outdoor scenes including mountains, cities, towns and urban sceneries. Images in this dataset are of different scales and views, which enhances the diversity across scenes. Second, images are selected elaborately to ensure they exhibit various appearances of the same scene. Moreover, the authors of \cite{Laffont14} defined 40 transient attributes for this dataset including weather related attributes (e.g., 'sunny', 'rain', 'fog', etc.). For each image, the weather related attributes are annotated non-exclusively, which is important for our multi-label weather recognition experiments. Several examples of the transient attribute dataset are illustrated in Figure \ref{transient_example}.

For weather recognition, six weather related attributes among all 40 transient attributes are selected, i.e., 'sunny', 'cloudy', 'fog', 'snow', 'moist' and 'rain', others are ignored in our experiments. Besides, we find that there exists a few examples in which all weather attribute strengths are very low. Some of them are captured at dawn and dusk, others do not show obvious features corresponding to any weather categories. Therefore, we add an 'other' class to represent those examples where every attribute strength is lower than 0.5. It is noteworthy that the strength lower than 0.5 indicates the annotation workers do not think the image exhibits the corresponding attribute. In this paper, for the weather recognition task, weather attributes greater and lower than 0.5 are set to 1 and 0, respectively. Finally, the dataset contains seven weather classes and 8571 images in total. The detailed statistics of the dataset are displayed in Table \ref{dataset_statistics}.

\subsubsection{The Multi-Label Weather Classification Dataset}
\begin{figure*}
		\centering
		\includegraphics[width=\textwidth]{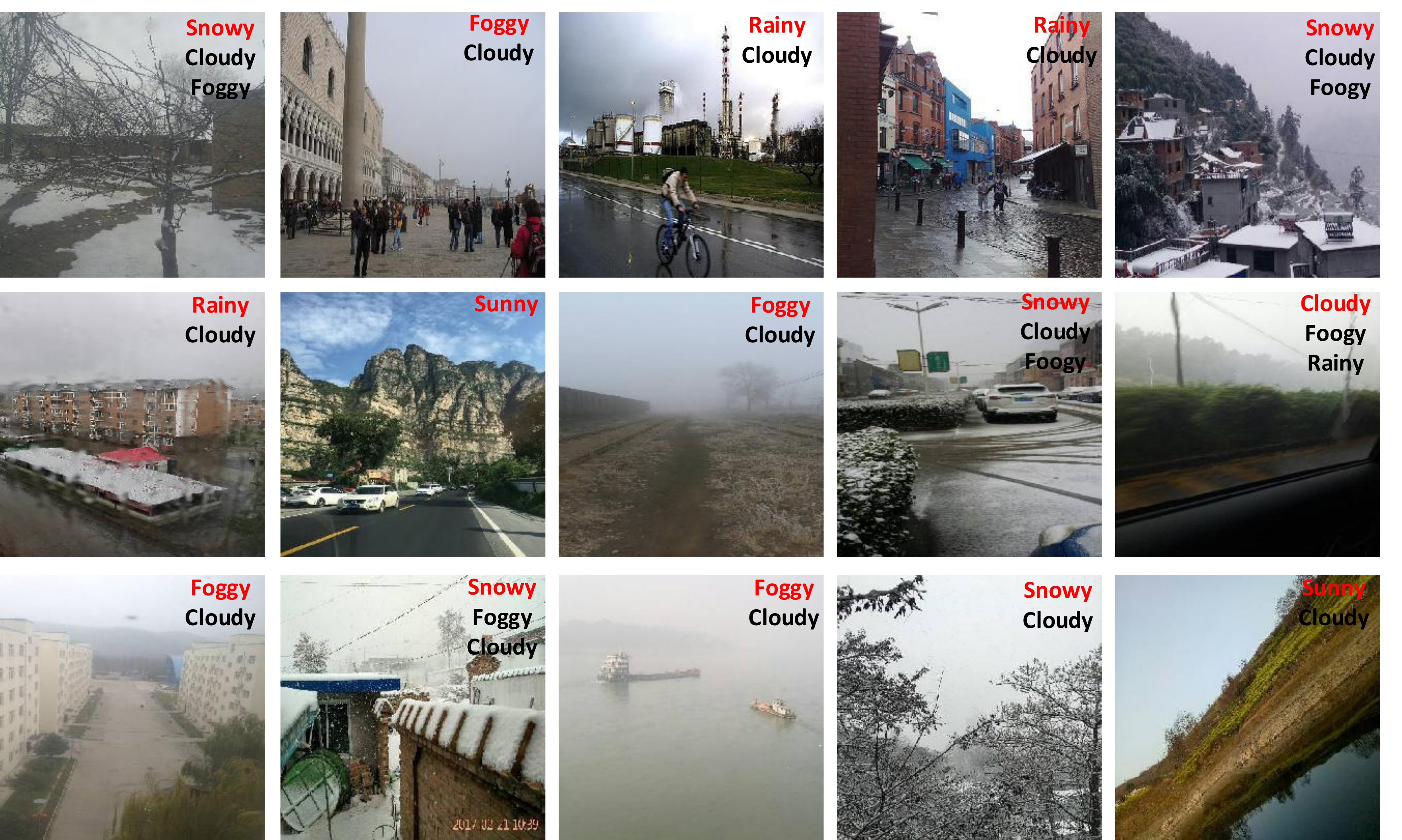}
		\caption{The illustration of examples in the multi-label weather classification dataset. The upper-right of each image displays the studied weather labels. Note that the red label denotes the weather class with the maximum strength.}
		\label{multilabel10000}
	\end{figure*}

To further evaluate the proposed approach, we construct a new dataset from scratch, which contains 10000 images from 5 weather classes, i.e., 'sunny', 'cloudy', 'foggy', 'rainy' and 'snowy'. All images are elaborately selected from Internet. Compared to other weather recognition datasets, our dataset has the following advantages. First, most of the existing datasets focus on only two or three weather classes, while our dataset covers all common weather conditions in the daily life. Second, the new constructed dataset contains many different scenes including cities, villages, urban areas and so on, as depicted in Figure \ref{multilabel10000}. In addition, this dataset also exhibits different scales and views. Third, in our dataset, the weather labels are not mutually exclusive, which can provide more weather information.  

The annotation of multi-label weather classification dataset was completed by a crowd-sourced task. The annotation workers are asked to determine weather attribute strengths non-exclusively for a given image, and the range of strengths is from 0 to 1, in which 0.5 is a demarcation point. Weather attribute strength lower than 0.5 indicates that the image cannot be judged as the corresponding weather condition (even if the image contains corresponding attribute). In this dataset, an image is annotated by at least five workers, and the average value of each attribute strength is selected as the result. To ensure the effectiveness of the annotation task, we also calculate the variance of each attribute strength for a given image. If the variance is bigger than a threshold, the result will be re-determined by discussion. Finally, to generate the weather labels, all attribute strengths greater than or equal to 0.5 are set to 1, others are set to 0. 

\begin{figure}
	\centering
	\subfigure[]{
		\includegraphics[width=5cm]{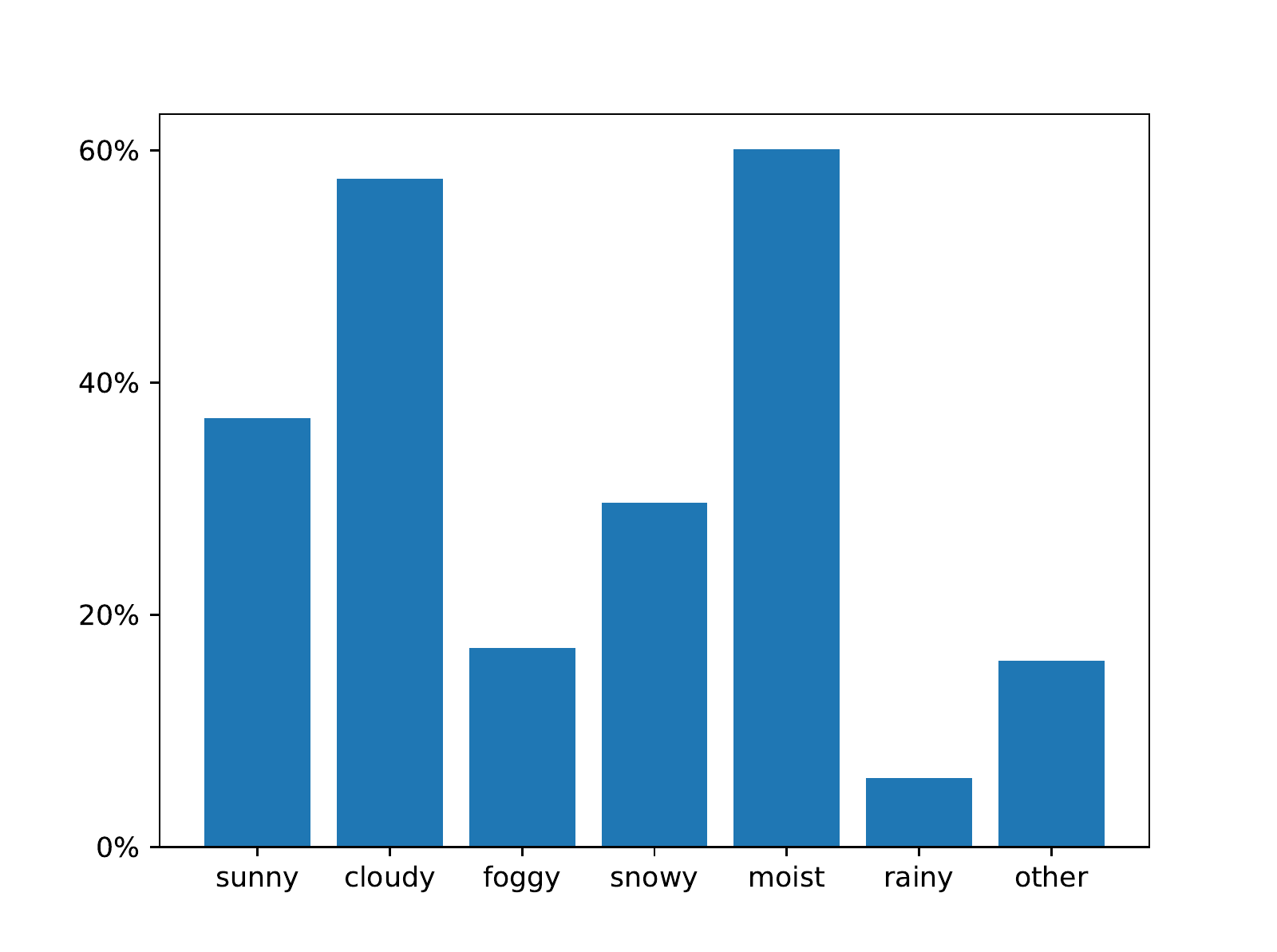}}
	\subfigure[]{
		\includegraphics[width=5cm]{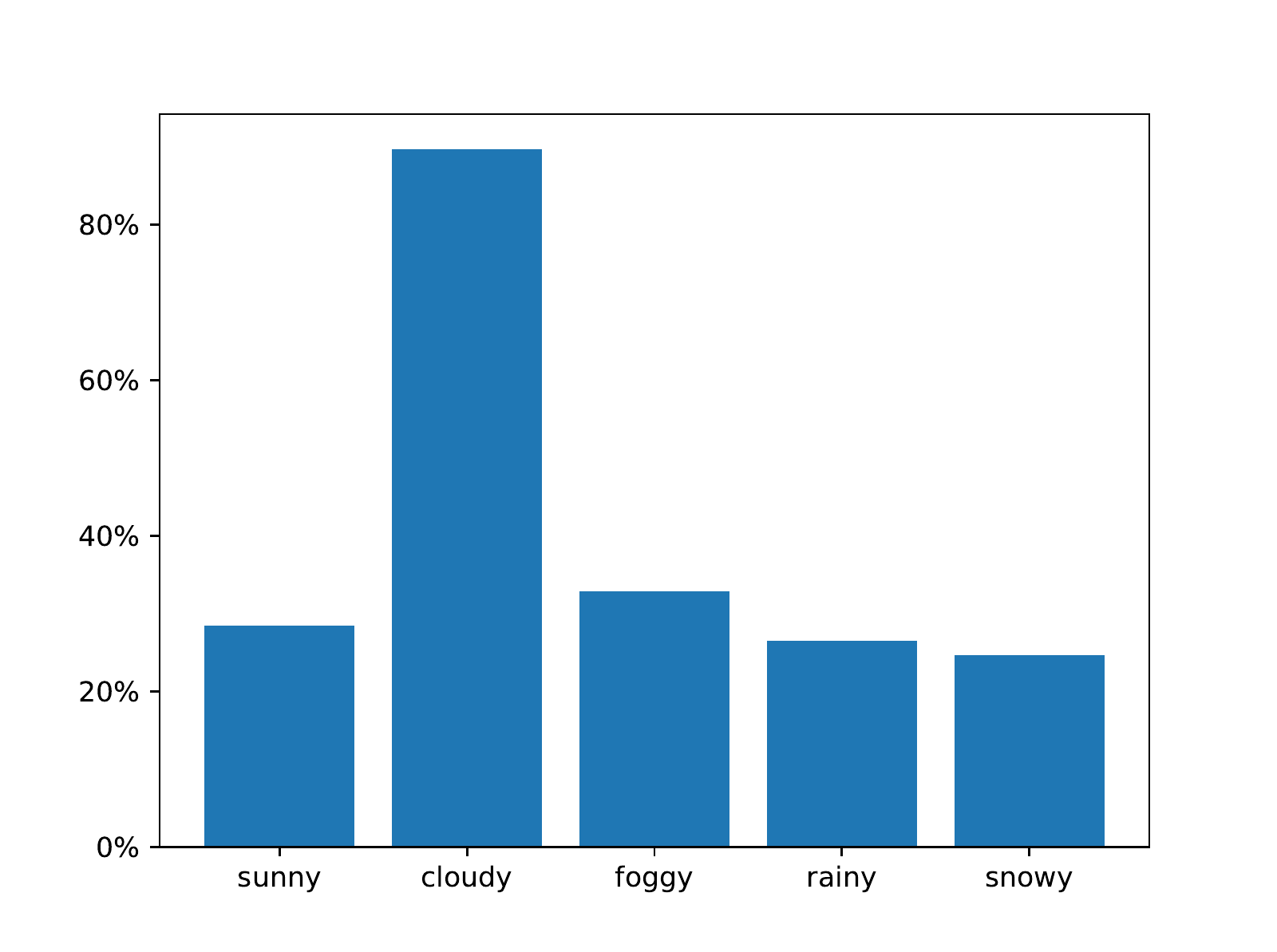}}
	\caption{Weather label distributions on two datasets (a) for transient attribute dataset and (b) for multi-label weather classification dataset.}
	\label{label_sta}
\end{figure}

Figure \ref{label_sta} shows the weather label distribution on the two experiment datasets. The detailed statistics can also be found at Table \ref{dataset_statistics}. In both datasets, cloudy is the class with large number of samples. This is because that cloudy usually co-occurs with other weather conditions. Apart from cloudy, the new constructed dataset is more balanced than the transient attribute dataset. Besides, it can be observed from Table \ref{dataset_statistics} that over half samples have multiple weather labels in both of the two datasets, which also verifies the validity of taking weather recognition as a multi-label classification task.

\begin{table*}
	\centering
	\caption{The statistics of the constructed datasets.}
	\resizebox{\linewidth}{!}{\begin{tabular}{|c|c|c|c|c|c|c|c|c|c|} 
			\hline
			Datasets   & Sunny & Cloudy  &Foggy &Snowy &Moist &Rainy &Other &$>$1 label&Total\\ \hline
			Transient attribute dataset & 2637 & 4113  &1224 &2120 &4295 &424 &1148 &4471&8571\\ \hline
			Multi-label weather dataset & 2370 &7474  &2737 &2052 &-- &2212 &-- &5763&10000\\ \hline			
		\end{tabular}}
		\label{dataset_statistics}
	\end{table*}

\subsection{Co-occurrence Relationships} \label{Co-occurrence Relationships}
We have qualitatively argued that more than one weather conditions may occur simultaneously in one image. The quantitative analysis of co-occurrence relationships among different weather conditions is also conducted according to the following equation,
\begin{equation}
R(i,j) = \frac{{\sum\limits_\Omega  {conc(i,j)} }}{{\sum\limits_\Omega  {I(i)} }},
\label{relation}
\end{equation}
where both $i$ and $j$ denote a kind of weather condition, $R(i,j)$ indicates the measurement of the co-occurrence relationship between $i$ and $j$. $\Omega$ represents all the samples in the dataset. $conc(i,j)$ and $I(i)$ are indicator functions which are defined as follows,
\begin{equation}
conc(i,j) = \left\{ {\begin{array}{*{20}{c}}
	{1,Arr(i) \ge 0.5 \wedge Arr(j) \ge 0.5}\\
	{0,\, \quad \quad \quad \quad \quad \quad \quad otherweise}
	\end{array}} \right.,
\end{equation}

\begin{equation}
I(i) = \left\{ {\begin{array}{*{20}{c}}
	{1,Arr(i) \ge 0.5}\\
	{0,\;\, otherweise}
	\end{array}} \right.,
\end{equation}
where $Arr(i)$ denotes the attribute strength of weather condition $i$, $\wedge$ represents the conjunction symbols. In summary, Eq. \ref{relation} indicates the ratio between co-occurrence number of the two weather conditions and the occurrence number of weather condition $i$ over all images. Therefore, \(\sum\nolimits_j {R\left( {i,j} \right)} \) and \(\sum\nolimits_j {R\left( {j,i} \right)} \) represent the influence and dependence of label \(i\) to others, respectively. To exploit the dependencies when predicting the weather labels, it is natural for us to predict the most influential label first and the dependent label last. Based on this, the following equation is utilized to rank the weather labels,
\begin{equation}r = \frac{{\sum\nolimits_j {R\left( {i,j} \right)} }}{{\sum\nolimits_j {R\left( {j,i} \right)} }}.\end{equation}
Obviously, the label with a higher score of \(r\) should rank first.

\begin{figure}
	\centering
	\subfigure[]{
		\includegraphics[width=5cm]{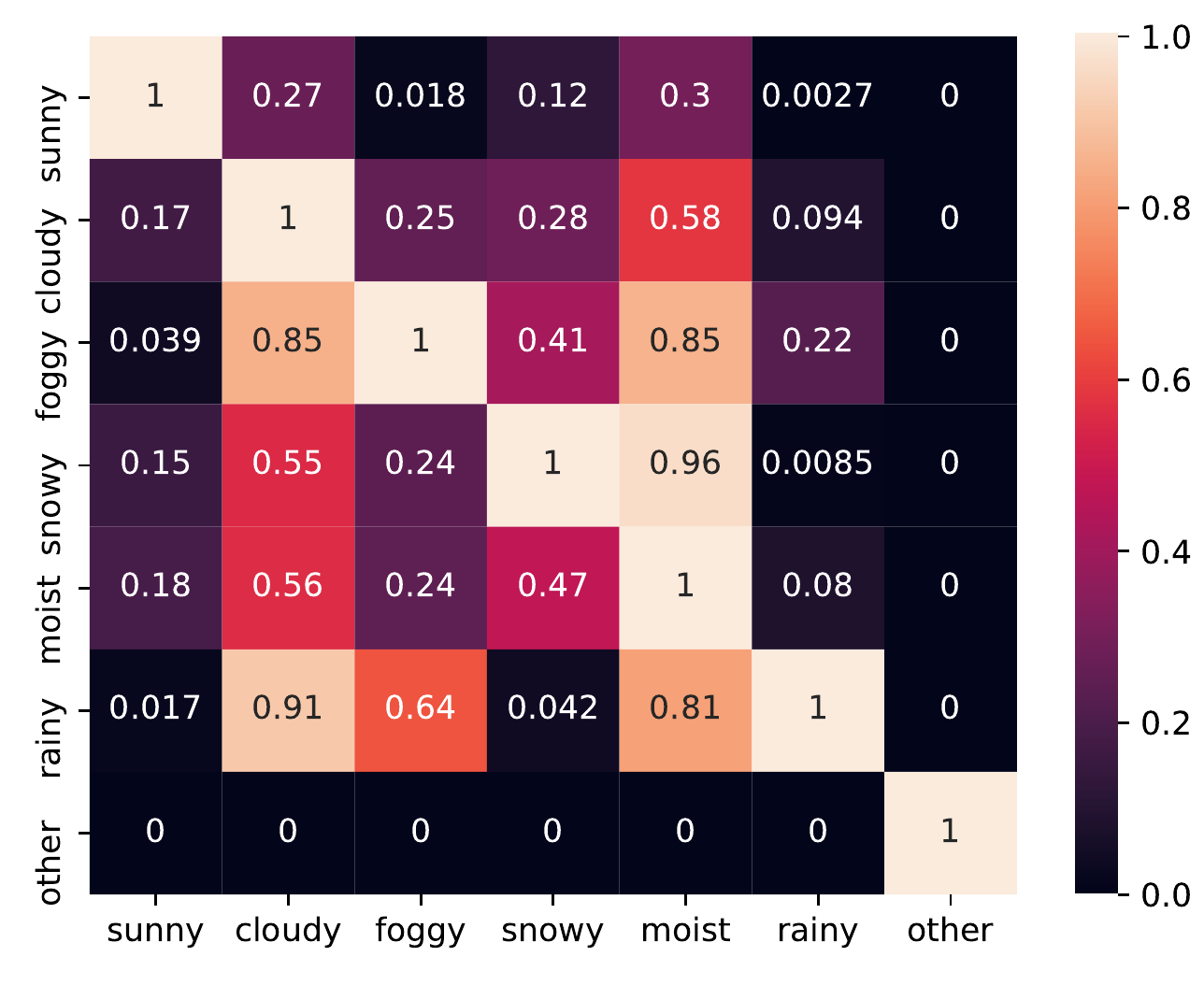}}
	\subfigure[]{
		\includegraphics[width=5cm]{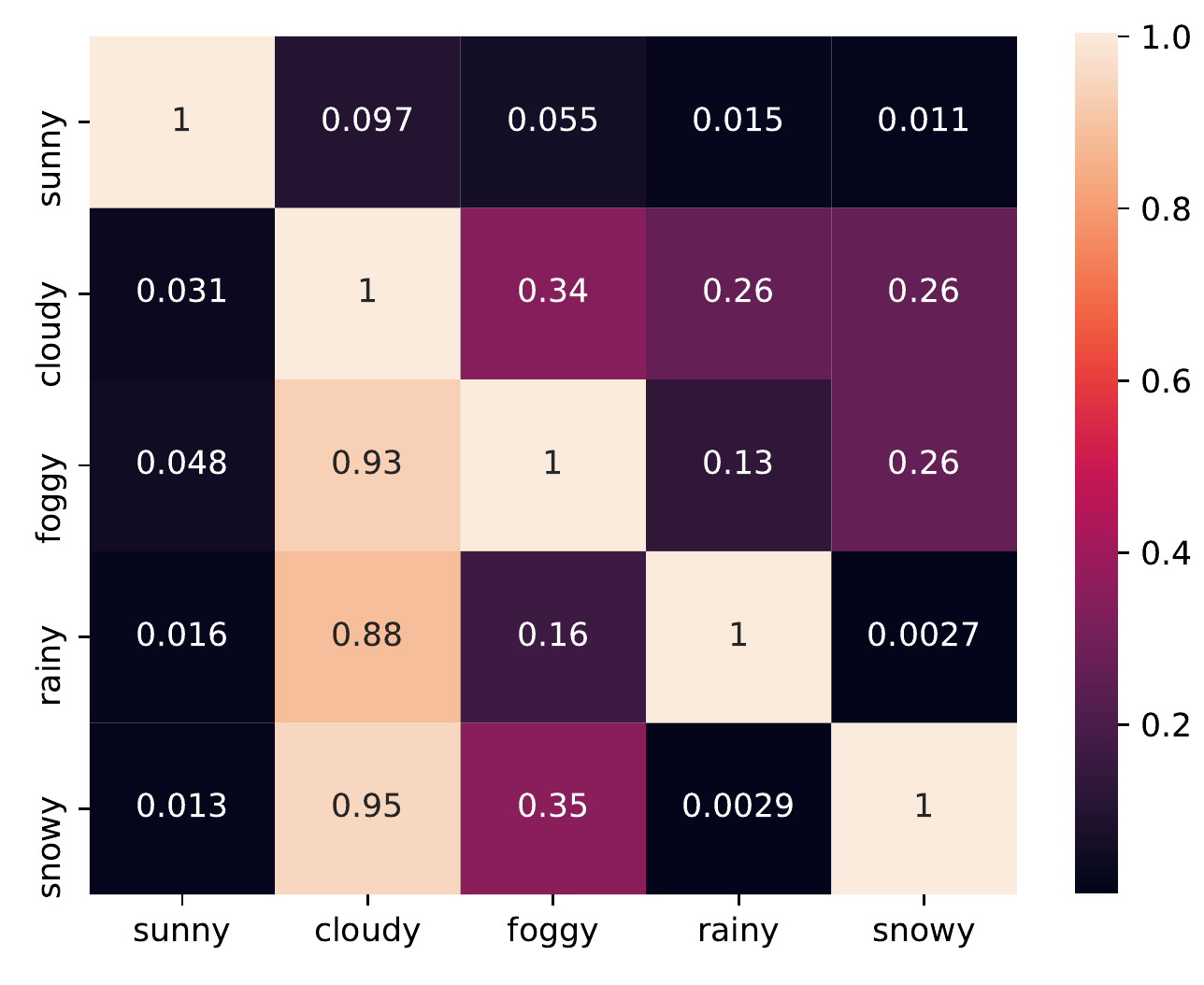}}
	\caption{The illustration of co-occurrence relationships among different weather conditions (a) for transient attribute dataset and (b) for multi-label weather classification dataset.}
	\label{co_occur}
\end{figure}

The analytical result is depicted in Figure \ref{co_occur}, from which we can simply draw the following conclusions. First, in accordance with our intuition, there are stronger co-occurrence relationships among different weather conditions, such as rainy and cloudy, snowy and foggy, etc. The corresponding samples are usually near the category boundary. In this paper, we propose to use the combination of labels to represent these samples. Second, there are indeed label dependencies in the weather recognition task. It is necessary to consider this problem when predicting multiple weather labels. In this paper, the convolutional LSTM is employed to capture the dependencies among different weather labels, and the labels are predicted step by step. According to Eq. 13, the order of weather labels is fixed as moist $\rightarrow$ cloudy $\rightarrow$ others $\rightarrow$ sunny $\rightarrow$ snowy $\rightarrow$ foggy $\rightarrow$ rainy on the transient attribute dataset, and cloudy $\rightarrow$ sunny $\rightarrow$ foggy $\rightarrow$ rainy $\rightarrow$ snowy on our multi-label weather classification dataset. Practically, we have also tried several other label orders, they get comparable performance, and the above two achieve the best in most occasions.

\subsection{Evaluation Metrics and Comparison Approaches}
Per-class precision and recall are first computed as evaluation metrics. Per-class means that for a given weather label, the prediction result is true as long as the current label is correctly predicted. Then, the average precision ($AP$) and average recall ($AR$) are calculated, which are defined as the average values of per-class precision and recall, respectively. 

Besides, sample-wise evaluation metrics are also adopted, which are defined as overall precision ($OP$) and overall recall ($OR$).
\begin{equation}
OP = \frac{{\sum\limits_{n = 1}^N {\sum\limits_{i = 1}^K f ({p_{n,i}},{{\tilde p}_{n,i}})} }}{{N \cdot K}},
\end{equation}
\begin{equation}
OR=\frac{\sum_{n=1}^N \sum_{i=1}^K f(p_{n,i},\widetilde{p}_{n,i})}{\sum_{n=1}^N \sum_{i=1}^K p_{n,i}},
\end{equation}
where $N$ denotes the number of samples in the dataset, $K$ represents the number of weather classes, $p_{n,i}$ and $\widetilde{p}_{n,i}$ indicate the ground-truth label and predicted label of the $n$-th sample on the $i$-th weather class, respectively. \(f\left(  \cdot  \right)\) is an indicator function which is defined as follows,
\begin{equation}
f(p,\widetilde{p}) = \left\{ {\begin{array}{*{20}{c}}
	{1,\quad \quad \; p=\widetilde{p}}\\
	{0,otherwise}
	\end{array}} \right..
\end{equation} 
Finally, the F1 scores (including AF1 and OF1) are computed, which are the harmonic mean of precision and recall.

Since there are no other multi-label weather recognition approaches, we compare with the multi-label version of AlexNet \cite{NIPS2012_4824} and VGGNet \cite{Simonyan14c}. To verify the effectiveness of convolutional LSTM and channel-wise attention model in this paper, we also compare with some other CNN-RNN frameworks, including CNN-LSTM, CNN-LSTM with spatial attention model (CLA), CNN-GRU with spatial attention model (CGA), CNN-ConvLSTM without channel-wise attention model. Besides, two widely used general multi-label approaches are also employed as comparison methods, i.e., ML-KNN \cite{ZHANG20072038} and ML-ARAM \cite{DBLP:conf/icdm/BenitesS15}. ML-KNN proposed a multi-label lazy learning method that adapts the traditional K-nearest neighbor (KNN) algorithm to the multi-label purpose. ML-ARAM extended the Adaptive Resonance Associative Map neural network for multi-label classification tasks. In our experiment, we test these two approaches using the implementations of the popular scikit-multilearn library. For fair comparisons, all CNN-RNN frameworks use the same CNN (i.e., VGGNet) with our approach. Features input to ML-KNN and ML-ARAM are also extracted by VGGNet (the last fully connection layer) pre-trained on two experimental datasets. The proposed approach are referred to as CNN-Att-ConvLSTM. 

\subsection{Results on the Transient Attribute Dataset}

\begin{table*}
	\centering
	\caption{The experimental result on the transient attribute dataset. (Per-class result: precision/recall)}
	\fontsize{6}{10}\selectfont
	\resizebox{\linewidth}{!}{\begin{tabular}{|c|c|c|c|c|c|c|c|c|c|c|c|c|c|}  
			\hline
			Approach   &Sunny &Cloudy &Foggy &Snowy &Moist &Rainy &Other &AP &AR &AF1 &OP &OR &OF1 \\ \hline
			
			AlexNet \cite{NIPS2012_4824} &0.756/0.892 &0.802/0.868 &0.688/0.688 &0.948/0.803 &0.840/0.903 &0.625/0.392 &0.789/0.224 &0.7783 &0.6815 &0.7267 &0.8967 &0.7999 &0.8455 \\ \hline
			
			VGGNet \cite{Simonyan14c} &0.777/0.836 &0.847/0.803 &0.767/0.717 &0.848/0.920 &0.873/0.899 &0.880/0.431 &0.622/0.552 &0.8022 &0.7369 &0.7682 &0.9043 &0.8155 &0.8576 \\
			\hline
			
			CNN-LSTM &0.819/0.754 &0.883/0.555 &0.777/0.529 &0.654/0.205 &0.986/0.142 &0.271/0.373 &0.000/0.000 &0.6271 &0.3653 &0.4617 &0.7991 &0.3814 &0.5163 \\ \hline
			
			CLA &0.856/0.816 &0.882/0.786 &0.861/0.63 &0.945/0.9 &0.883/0.913 &0.596/0.608 &0.547/0.657 &0.7958 &0.7585 &\textbf{0.7767	} &0.9117 &0.815 &0.8606 \\ \hline
			
			CGA &0.765/0.895 &0.861/0.805 &0.879/0.63 &0.949/0.892 &0.891/0.901 &0.632/0.471 &0.573/0.56 &0.7925 &0.7362 &0.7633 &0.9093 &0.8176 &0.861 \\ \hline
			
			CNN-ConvLSTM &0.868/0.777 &0.876/0.813 &0.789/0.703 &0.938/0.916 &0.867/0.929 &0.653/0.627 &0.548/0.552 &0.7913 &\textbf{0.7596} &0.7751 &0.912 &0.8203 &0.8637 \\ \hline
			
			ML-KNN \cite{ZHANG20072038} &0.720/0.892 &0.898/0.742 &0.866/0.609 &0.887/0.944 &0.818/0.933 &0.700/0.412 &0.663/0.425 &0.7927 &0.708 &0.7479 &0.9001 &0.8037 &0.8492 \\ \hline
			
			ML-ARAM \cite{DBLP:conf/icdm/BenitesS15} &0.790/0.826 &0.784/0.853 &0.903/0.609 &0.968/0.855 &0.938/0.842 &0.889/0.314 &0.550/0.739 &\textbf{0.8318} &0.7197 &0.7717 &0.9044 &0.8047 &0.8517 \\ \hline

			CNN-Att-ConvLSTM &0.857/0.785 &0.851/0.852 &0.837/0.682 &0.952/0.896 &0.913/0.911 &0.656/0.454 &0.585/0.628 &0.8091 &0.7428 &0.776 &\textbf{0.9167} &\textbf{0.8231} &\textbf{0.8678} \\ \hline
			
		\end{tabular}}
		\label{res_table1}
	\end{table*}

For the transient attribute dataset, 1000 images are randomly selected for testing, another 1000 images are selected for validation, and the remains are for training. The experimental result is shown in Table \ref{res_table1}, from which we can see that the proposed approach CNN-Att-ConvLSTM achieves the best results on OP, OR and OF1, and comparable results with the state-of-the-arts on AP, AR and AF1. CNN-LSTM with spatial attention model (CLA) also gets good results. While without spatial attention model, CNN-LSTM suffers from serious performance degradation. This indicates the importance of some key regions in the weather recognition task. To evaluate the influences of LSTM in the CNN-RNN framework, we also test CNN-GRU with spatial attention model (CGA), and find CGA achieves almost the same results with CLA. CNN-ConvLSTM also gets similar results with CLA, which denotes the effectiveness of convolutional LSTM in information extraction of key regions. Overall, the proposed approach perform better than multi-label version of AlexNet, VGGNet, the general multi-label approaches ML-KNN, ML-ARAM, and other CNN-RNN methods, which proves the superiority of our approach. 

For per-class result, all these methods perform worse on 'rainy' and 'other' classes. This is because that most images in transient attribute dataset present distant views. It is difficult to recognize the rainy weather from such distant views. In addition, samples of 'other' class are very rare, and can be easily misclassified as sunny or cloudy in this dataset.

	\begin{table*}
		\centering
		\caption{The experimental result on multi-label weather classification dataset. (Per-class result: precision/recall)}
		\fontsize{6}{10}\selectfont
		\resizebox{\linewidth}{!}{\begin{tabular}{|c|c|c|c|c|c|c|c|c|c|c|c|}  
				\hline
				Approach   &Sunny &Cloudy &Foggy &Rainy &Snowy &AP &AR &AF1 &OP &OR &OF1 \\ \hline
				
				AlexNet \cite{NIPS2012_4824} &0.84/0.74 &0.896/0.942 &0.735/0.89 &0.784/0.685 &0.876/0.905 &0.8263 &0.8325 &0.8294 &0.9007 &0.8668 &0.8834 \\ \hline
				
				VGGNet \cite{Simonyan14c} &0.772/0.851 &0.927/0.915 &0.867/0.728 &0.814/0.701 &0.887/0.931 &0.8533 &0.8252 &0.839 &0.9087 &0.8494 &0.878 \\ \hline
				
				CNN-LSTM &0.843/0.791 &0.897/0.958 &0.86/0.73 &0.83/0.694 &0.94/0.556 &\textbf{0.8739} &0.7458 &0.8048 &0.8991 &0.8127 &0.8537 \\ \hline
				
				CLA &0.853/0.791 &0.908/0.944 &0.863/0.772 &0.856/0.687 &0.924/0.924 &0.8809 &0.8237 &0.8513 &0.9169 &0.8582 &0.8866 \\ \hline
				
				CGA &0.858/0.785 &0.901/0.954 &0.825/0.802 &0.829/0.729 &0.924/0.924 &0.8671 &0.8387 &0.8527 &0.9161 &0.8722 &0.8936 \\ \hline
				
				CNN-ConvLSTM &0.855/0.78 &0.899/0.953 &0.798/0.862 &0.843/0.716 &0.926/0.924 &0.8643 &0.8472 &0.8557 &0.9165 &0.8793 &0.8975 \\ \hline
				
				ML-KNN \cite{ZHANG20072038} &0.837/0.724 &0.912/0.94 &0.819/0.834 &0.794/0.736 &0.918/0.934 &0.8562 &0.8336 &0.8447 &0.9138 &0.8766 &0.8948 \\ \hline
				
				ML-ARAM \cite{DBLP:conf/icdm/BenitesS15} &0.772/0.81 &0.853/0.936 &0.952/0.783 &0.641/0.762 &0.979/0.84 &0.8397 &0.8262 &0.833 &0.8988 &0.865 &0.8816 \\ \hline
				
				CNN-Att-ConvLSTM &0.838/0.843 &0.917/0.953 &0.856/0.861 &0.856/0.758 &0.894/0.938 &0.8721 &\textbf{0.8702} &\textbf{0.8705} &\textbf{0.9263} &\textbf{0.8946} &\textbf{0.9135} \\ \hline
				
			\end{tabular}}
			\label{res_table2}
		\end{table*}

\subsection{Results on the Multi-label Weather Classification Dataset}

For multi-label weather classification dataset, 2000 images are randomly selected for testing, 1000 images for validation, and the remains for training. As presented in Table \ref{res_table2}, CNN-Att-ConvLSTM performs the best on almost all the evaluation metrics, which demonstrates the effectiveness of the proposed approach again.

\begin{figure*}
		\centering
		\includegraphics[width=1\textwidth]{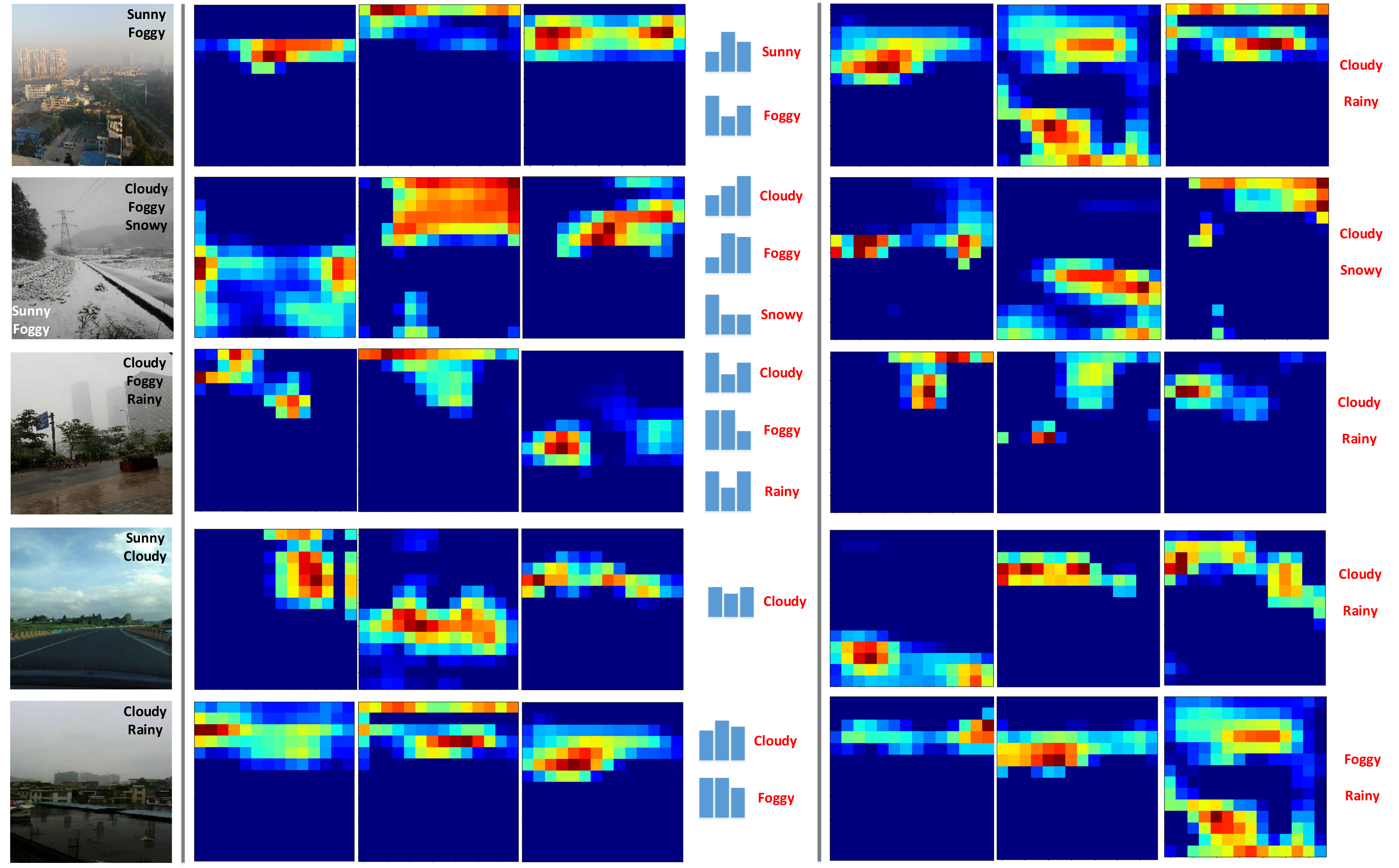}
		\caption{Five results from the multi-label weather classification dataset. There are three columns: 1) The images and their annotated labels. 2) The representative activation maps from CNN-Att-ConvLSTM and their attention weights (blue histograms), together with the classification results. 3) The representative activation maps from the baseline (VGGNet-based multi-label classification) and the classification results.}
		\label{activation}
	\end{figure*}

To analyze the effectiveness of our approach, some weather recognition examples are presented in Figure \ref{activation}. It includes the classification results, activation maps and attention weights from our approach. The results of VGGNet are utilized for comparison, since our approach also uses it as the deep feature extractor. 
Specifically, our approach works well on the above three images. From the selected activation maps and their attention weights, we can see that our approach can attend to the most correlated weather cues when predicting different weather labels, while the results of VGGNet are not so satisfactory. For example, the first image is annotated as sunny and foggy, correspondingly the blue sky, the bright area and the region of haze have stronger responses in our activation maps, and the attention weights of corresponding activation maps are relatively high when predicting different labels. However, the ground is activated by VGGNet mistakenly, which leads to the wrong label, i.e., rainy. Besides, our approach fails on the rest two images, where the fourth image is annotated as sunny and cloudy, which means an intermediate state between sunny and cloudy. However, only the cloud regions are activated, and the sunny label is lost in our approach. It is mainly because the sunny label is a little ambiguous. The fifth image is annotated as cloudy and rainy. However, due to the wet ground is not so obvious, it is mis-classified as cloudy and foggy in our approach. Overall, the results in Figure \ref{activation} indicate that our approach performs well in most cases, but sometimes fails when the annotation is ambiguous and the weather cues are not obvious. It is reasonable since our approach is just based on the visual features, and maybe better performance can be achieved with other modality information, such as humidity, which can be taken into consideration in our future work.

\section{Conclusion}\label{conclusion}
Considering that more than one weather conditions may occur simultaneously in one image, we firstly analyze the drawbacks of taking weather recognition as a single label classification task, and propose a multi-label classification framework for the weather recognition task. It allows one image to belong to multiple weather categories, which can provide more comprehensive description of the weather conditions. Specifically, it is a CNN-RNN architecture, where CNN is extended with a channel-wise attention model to extract the most correlated visual features, and a convolutional LSTM is utilized to predict the weather labels step by step, meanwhile, maintaining the spatial information of the visual feature. Besides, we build two datasets for the weather recognition task to make up the problem of lacking training data. Practically, the experimental results have verified the effectiveness of the proposed approach.  

In the future work, we plan to introduce the distribution prediction task for weather recognition \cite{DBLP:conf/ijcai/ZhaoDGH17,DBLP:conf/mm/ZhaoDGH17,zhao2018discrete,zhao2017continuous}, which can not only classify the image with multi-labels, but also predict the strengths of different weather class, so as to describe the weather conditions more comprehensively. Besides, other modality information, such as humidity and temperature, can also be utilized in the future work.

\section*{References}

\bibliography{elsarticle-template}

\end{document}